\documentclass[journal]{IEEEtran}
\ifdefined\pdfshellescape\else
  \newcount\pdfshellescape
\fi
\usepackage{amsmath}
\usepackage{stfloats} 
\usepackage[pdftex]{graphicx}
\usepackage{caption}
\usepackage{subfiles} 
\usepackage{subfigure}
\usepackage{graphicx}

\usepackage{enumitem}
\usepackage{amsmath,amssymb,amsopn,amstext,amsfonts}
\usepackage{newtxtext,newtxmath}
\usepackage{cancel}
\usepackage[space]{cite}
\usepackage{balance}
\usepackage{color}
\usepackage[table]{xcolor}
\usepackage{algorithm}
\usepackage{algorithmic}
\usepackage{url}
\usepackage{booktabs}
\usepackage{threeparttable}
\usepackage[linkcolor=black,citecolor=black,urlcolor=black,colorlinks=true]{hyperref}
\usepackage{multirow}
\usepackage[table]{xcolor} 
\usepackage{colortbl}
\usepackage{graphicx}
\usepackage{array}
\usepackage{verbatim}
\usepackage{makecell}
\usepackage{stackengine}
\usepackage{balance}
\usepackage{paracol}
\definecolor{mygreen}{RGB}{219, 234, 192}

\graphicspath{{./figure/}}

\DeclareGraphicsExtensions{.pdf,.png,.jpg,.PNG}
\title{TravExplorer: Cross-Floor Embodied Exploration via Traversability-Aware 3-D Planning}
\author{
Han Zheng$^1$, Zhe Chen$^1$, Yudong Huang$^1$, Haoran Liu$^1$, Jinghao Wang$^1$, Ming Yang$^1$, and Tong Qin$^{1\ast}$
    \thanks{
        $^{1}$Shanghai Jiao Tong University, Shanghai, China.
        {\{hanzheng, qintong\}@sjtu.edu.cn}.
    }
    \thanks{
    {$^\ast$ is the Corresponding author. This work was supported by the Natural Science Foundation of Shanghai (Grant No. 24ZR1435600).}
    }

}

\begin{document}
\maketitle


\begin{abstract}
Zero-shot Object Navigation (ZSON) has shown promise for open-vocabulary target search in unseen environments, yet most existing systems remain tied to planar representations and single-floor assumptions. These assumptions become
inadequate in real buildings, where navigation involves floors,
stairs, landings, and vertically overlapping spaces. This article presents TravExplorer, a cross-floor embodied exploration framework that couples zero-shot semantic guidance with traversability-aware 3-D planning. TravExplorer maintains a unified volumetric map that distinguishes occupied structures from robot-reachable support surfaces and extracts traversable frontiers from connected support surfaces, including floors, stairs, and landings. A FOV-aware active perception strategy further resolves incomplete observations during cross-floor traversal. To reduce semantic-reasoning latency, a lightweight guidance module aligns a probabilistic instance map from online open-vocabulary segmentation with a spatial value map from fast image-to-text matching. Based on these geometric and semantic memories, a hierarchical planner performs target-aware frontier touring over object hypotheses, traversable frontiers, and stair landmarks, and generates executable cross-floor motions through foothold-guided 3-D search and vertically constrained local trajectory optimization. Experiments over 4,195 simulated episodes on HM3D and MP3D demonstrate consistent advantages over representative ObjectNav baselines. Fifty real-world trials on a Unitree Go2 further validates open-vocabulary target search across single-floor and cross-floor indoor environments without prior maps or human intervention.
The code will be released at \href{https://github.com/wuyi2121/TravExplorer}{\textcolor{blue}{https://github.com/wuyi2121/TravExplorer}}.
\end{abstract}


\section{Introduction}

\begin{figure}[t]
	\centering
	\includegraphics[width=1.0\linewidth]{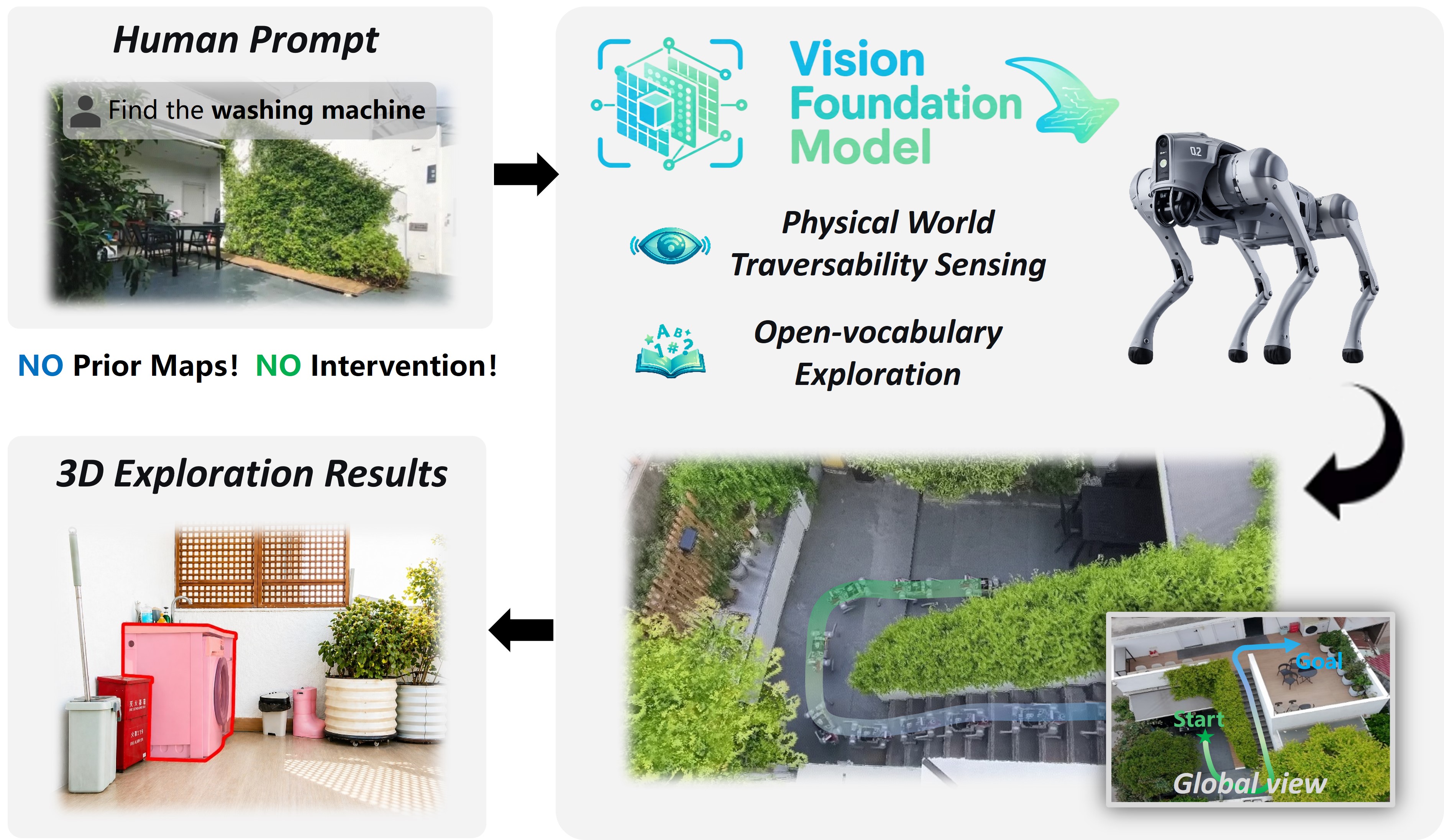}
    \captionsetup{font=footnotesize} 
	\caption{Motivating scenario of TravExplorer.
Given an open-vocabulary object query, the quadruped robot explores an unknown building without prior maps or human intervention.
TravExplorer couples foundation-model perception with 3-D traversability reasoning, enabling the robot to search across floors, stairs, and landings while reaching executable viewpoints for target localization.}
	\label{fig:teaser}
\end{figure}

Object navigation (ObjectNav) requires a robot to search for a target object in an unseen environment and stop near a valid instance. 
In real service and inspection scenarios, a simple query such as ``find the fire extinguisher'' may require the robot to explore rooms, corridors, stairs, and landings while reasoning from partial observations. 
Recent large language models (LLMs) and vision-language models (VLMs) have made zero-shot ObjectNav (ZSON) increasingly practical by enabling open-vocabulary perception and commonsense-guided exploration~\cite{gadre2023cows,yu2023l3mvn,yokoyama2024vlfm,yin2024sgnav}. 
However, most existing systems are still developed under planar and single-floor assumptions. 
This mismatch becomes critical in real buildings, where the robot and the target may lie on different floors connected by stairs or other non-planar support surfaces.
Fig.~\ref{fig:teaser} illustrates the cross-floor search scenario considered in this work.

A common ZSON pipeline constructs a 2-D occupancy or semantic map, extracts frontiers on the boundary of known free space, and ranks candidate frontiers using VLM or LLM cues~\cite{yokoyama2024vlfm,zhang2025apexnav,zhu2025strive}. 
This paradigm has shown strong performance in planar environments, but two issues remain for long-horizon cross-floor search. 
First, local semantic scores from early or occluded views can be unstable, leading to redundant backtracking or false stopping~\cite{zhang2025apexnav,zhu2025strive}. 
Second, repeatedly querying large models for frontier or viewpoint selection introduces non-negligible latency and computation cost on robotic platforms~\cite{zhu2026sysnav,gai2026ussnav}. 
Cross-floor search amplifies both issues because semantic evidence must be accumulated and reused across larger, vertically connected spaces.

The geometric difficulty is more fundamental. 
For a ground or legged robot, floors, stairs, ramps, and landings are traversable support surfaces, although they appear as occupied geometry in a volumetric map. 
A pure occupancy representation may therefore remove the very surfaces on which the robot should move. 
In contrast, a top-down projection preserves planning efficiency but collapses vertically separated floors and breaks stair connectivity. 
Recent floor-aware ZSON methods introduce multi-floor reasoning~\cite{zhang2024mfnp,gong2026ascent}, yet their layer-wise maps and discrete floor transitions still fragment the underlying 3-D traversable space. 
Existing 3-D navigation frameworks can preserve multi-layer geometry~\cite{yang2024pct}, but they usually assume a known or sufficiently reconstructed map and do not address online semantic exploration.

These observations suggest that cross-floor ObjectNav is neither a semantic target-search problem nor a 3-D path-planning problem. 
The robot must discover traversable support surfaces online, maintain revisitable connectivity across floors, use semantic cues without excessive large-model calls, and convert exploration decisions into executable motion under limited field of view. 
In particular, stair regions require active perception, since incomplete observations may disconnect the very transition that enables cross-floor search.

To this end, we propose \textit{TravExplorer}, a cross-floor embodied exploration system for quadruped robots in 3-D traversable spaces. 
TravExplorer maintains a volumetric occupancy map with a semantic traversability layer, representing floors, stairs, and landings in a unified metric frame. 
It extracts frontiers directly on connected traversable support surfaces and uses active perception to complete uncertain stair observations. 
For semantic guidance, TravExplorer fuses online open-vocabulary segmentation, probabilistic instance memory, and lightweight image-text matching. 
For motion generation, a hierarchical planner couples target-aware frontier touring, foothold-guided 3-D graph search, and local trajectory optimization.

The main contributions are summarized as follows:

(i)
We pioneer a traversable frontier-based exploration paradigm for ground robots, which unifies 2D ray casting for efficiency and 3D semantic traversability for multi-floor awareness, while leveraging active perception to mitigate FOV-limited occlusions.

(ii)
We design spatial semantic guidance that couples a probabilistic instance map from open-vocabulary semantic segmentation with a spatial value map from fast image-to-text matching, enabling zero-shot search to focus on semantically promising regions for efficient exploration.

(iii)
We develop a hierarchical cross-floor exploration planner that integrates TSP-based global tours, foothold-guided 3D path search, an execute-review mechanism, and vertically constrained local planning, enhancing both search efficiency and traversal robustness.

(iv)
We conduct extensive evaluations, including 50 real-world experiments and 4,195 simulated episodes on HM3D and MP3D benchmarks, achieving state-of-the-art performance. To the best of our knowledge, this is the first system that efficiently handles both single-floor and cross-floor object navigation through 3D traversability mapping.

\section{Related Work}
\label{sec:Related Work}

\subsection{Autonomous Exploration.}

Autonomous exploration is commonly formulated as selecting informative viewpoints that expand the known map while keeping the motion executable. 
Frontier-based exploration~\cite{yamauchi1997frontier} identifies boundaries between known and unknown regions as exploration targets, providing a simple and scalable basis for mobile robots. 
Efficient dense frontier detection~\cite{orsulic2019frontier} accelerates frontier extraction on 2-D graph-SLAM submaps, FAEL~\cite{huang2023fael} improves large-scale mobile-robot exploration efficiency, and region partitioning~\cite{wen2025region} reduces redundant long-range transfers. 
These methods are effective for planar occupancy or projected maps, but the projection collapses stairs and vertically overlapping corridors into ambiguous 2-D cells.

For richer spatial coverage, another line of work performs exploration in 3-D volumetric spaces. 
FUEL~\cite{zhou2021fuel} maintains an incremental frontier information structure and uses hierarchical planning to generate fast UAV motions. 
FALCON~\cite{zhang2024falcon} further introduces connectivity-aware space decomposition and coverage-path guidance, aligning local frontier visitation with a global intention over unexplored space. 
3-D Active Metric-Semantic SLAM~\cite{tao2024active3d} incorporates localization uncertainty into active metric-semantic reconstruction, EPIC~\cite{geng2025epic} uses lightweight LiDAR point-cloud observation maps for large-scale UAV exploration, and FLARE~\cite{liu2025flare} partitions unknown regions to guide fast exploration. 
Although these explorers better preserve 3-D topology and reduce revisitation, their candidate viewpoints are generated in free space and are not required to lie on ground-reachable support surfaces.

Ground-robot exploration must additionally reason about terrain traversability, stability, and collision clearance. 
SMUG Planner~\cite{chen2023smug} performs safe multi-goal planning with robot-specific traversability constraints, LRAE~\cite{bi2024lrae} extracts large safe exploration regions on uneven terrain, and Portable Planner~\cite{jia2024portable} improves exploration robustness in unstructured environments. 
AAGE~\cite{zheng2025aage} uses aerial bird's-eye-view priors to assist ground robots in large unknown scenes. 
However, these systems are mainly designed for outdoor rough terrain or 2.5-D elevation-map settings. 
They do not explicitly address indoor multi-floor exploration, where different floors may overlap in the horizontal plane and vertical transitions must remain continuously connected to the traversable map.

\subsection{Cross-Floor Trajectory Planning.} 

Ground navigation beyond planar workspaces is often built on terrain representations that expose height and traversability. 
Probabilistic terrain mapping~\cite{fankhauser2018probabilistic} estimates grid-based elevation with uncertainty bounds under drifting proprioceptive localization, and GPU elevation mapping~\cite{miki2022elevation} accelerates smoothing, inpainting, and segmentation for locomotion and navigation. 
Perceptive locomotion through nonlinear MPC~\cite{grandia2023perceptive} integrates terrain perception into whole-body control, while uneven-terrain trajectory planning~\cite{xu2023uneven} uses terrain pose mapping to couple traversability assessment with car-like robot dynamics. 
Such 2.5-D maps are compact and effective for slopes, steps, and rough ground. 
However, they normally store one dominant surface per grid cell, which makes them unreliable in indoor buildings with overhangs, stair voids, and vertically overlapping floors.

To extend ground navigation to multi-layer structures, recent methods preprocess 3-D geometry into lower-dimensional but structured planning spaces. 
Beyond-2-D trajectory generation~\cite{wang2023beyond2d} constructs an \(R^3\) penalty field so that active and passive height variations can be optimized in a continuous trajectory space. 
PCT~\cite{yang2024pct} analyzes point clouds tomographically, generating slices that encode ground and ceiling elevations while evaluating traversability according to robot motion capabilities. 
Traversable Planes~\cite{zhang2025traversableplanes} exploits the fact that architectural spaces are often composed of ground, slopes, and stairs, and builds a lightweight plane graph through segmentation, merging, classification, and connection. 
These representations significantly reduce the complexity of 3-D search, yet they are mainly designed for goal-directed planning after sufficient geometric observations have been acquired.

Recent robot-specific frameworks further move from offline or goal-directed
planning toward online navigation and exploration in complex 3-D terrain.
Real-time multilevel terrain-aware planning~\cite{li2025multilevel}
introduces an implicit global map and local configuration-stability evaluation
for large-scale rough terrains, while TiFA~\cite{wang2025tifa} provides a
terrain-informed navigation framework for articulated tracked robots in rescue
missions. These methods show that online terrain reasoning is feasible beyond
prebuilt maps. However, they often rely on specialized robot morphologies,
task-specific terrain representations, or carefully designed traversability
segmentation, which limits their generality for cross-floor ObjectNav.

\subsection{Zero-Shot Object Navigation.} 

ObjectNav requires an agent to find a target object category in an unseen
environment, coupling semantic recognition with long-horizon exploration. 
Early modular methods, such as SemExp~\cite{chaplot2020object} and
PONI~\cite{ramakrishnan2022poni}, rely on semantic maps or learned object-location priors to guide exploration, while ZSON~\cite{majumdar2022zson} extends
ObjectNav to unseen categories by aligning language goals with visual
observations. 
More recently, foundation-model-based approaches have shifted
zero-shot ObjectNav toward open-vocabulary perception and explicit semantic
reasoning. 
Among them, VLFM~\cite{yokoyama2024vlfm} is particularly relevant to
our work, as it constructs language-grounded value maps over frontier candidates
to guide semantic exploration.
SG-Nav~\cite{yin2024sgnav} prompts LLMs with online 3-D scene graphs, FOM-Nav~\cite{chabal2025fomnav} organizes frontier-object maps for object-goal search, and OpenFrontier~\cite{padilla2026openfrontier} grounds frontiers with visual-language cues. 
ApexNav~\cite{zhang2025apexnav} adaptively switches between semantic reasoning and geometry-based exploration according to the distribution of semantic cues, while using target-centric semantic fusion to reduce false detections. 
STRIVE~\cite{zhu2025strive} builds a multi-layer representation of viewpoints, objects, and rooms, SysNav~\cite{zhu2026sysnav} decouples reasoning, planning, and control, and USS-Nav~\cite{gai2026ussnav} constructs a unified spatio-semantic scene graph for lightweight UAV navigation. 
These methods improve semantic guidance and reduce unnecessary model queries, but their navigation backbones are still largely based on 2-D frontiers, room-level abstractions, or embodiment-specific motion layers.

Multi-floor zero-shot ObjectNav is a more recent direction. 
MFNP~\cite{zhang2024mfnp} pioneers multi-floor ZSON by introducing a multi-floor navigation policy, multimodal LLM reasoning, and inter-floor navigation through stairs. 
However, its stair traversal mechanism resets the map after floor transition, discards previous exploration memory, and marks the stair entrance to enforce one-way movement. 
ASCENT~\cite{gong2026ascent} improves online floor-aware navigation with a multi-floor abstraction, stair-aware obstacle mapping, cross-floor topology modeling, and LLM-driven coarse-to-fine exploration. 
It maintains separate floor-wise maps and switches among them, alleviating memory loss but still breaking the continuity of the 3-D environment into discrete 2-D layers. 
Moreover, floor transitions remain dependent on predefined rules, limiting adaptation to diverse stair geometries. 
TravExplorer instead maintains a unified 3-D traversability map and couples zero-shot semantic guidance with continuous cross-floor motion planning.

\begin{figure*}[t]
	\centering
	\includegraphics[width=0.95\textwidth]{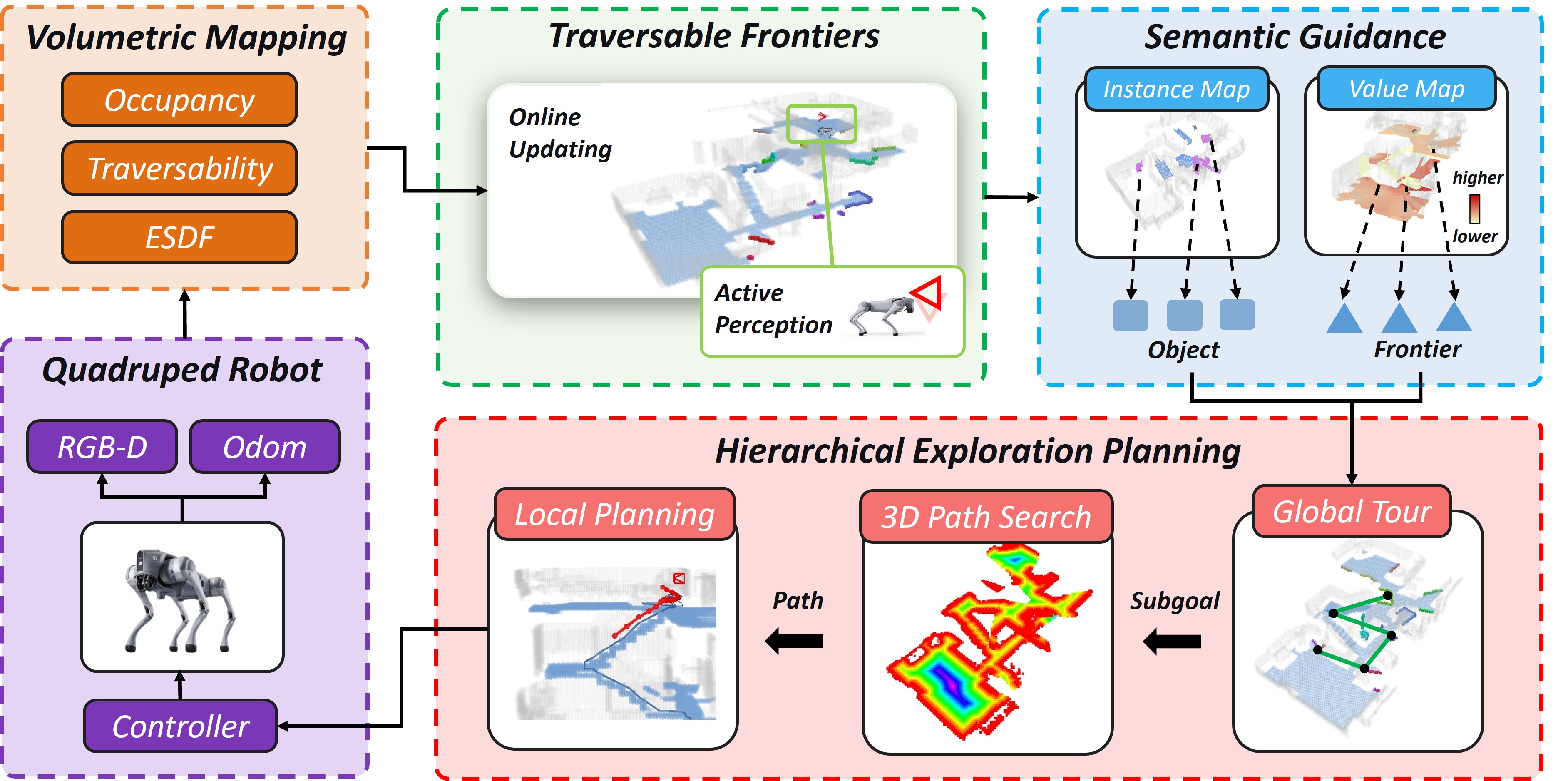}
	\captionsetup{font=footnotesize}
	\caption{System overview of TravExplorer.
The quadruped robot receives RGB-D and odometry observations and incrementally maintains volumetric occupancy, traversability, and ESDF layers.
Traversable frontiers are extracted on connected reachable support surfaces and updated online with active perception.
In parallel, the dual-semantic guidance module builds a target-centric instance map and a spatial value map to associate open-vocabulary object evidence with reachable frontier regions.
The hierarchical exploration planner then performs global frontier touring, traversability-aware 3-D path search, and local planning, closing the loop by sending executable commands to the robot controller.
}
	\label{fig:framework}
\end{figure*}

\section{System Overview}
\label{sec:Method}
\subsection{Problem Formulation}

We first define the task interface, observations, actions, and success criterion used throughout the paper.
The ObjectNav task \cite{batra2020objectnav} involves the agent navigating in a previously unseen environment to find an instance of the target object category (e.g., ‘chair’). 
The target-object query set is represented by \(Q_{\mathrm{obj}} = \{q_1, \ldots, q_n\}\), and the scene set is represented by \(\mathcal{S} = \{s_1, \ldots, s_m\}\). 
Each episode begins with the agent being initialized at state \(\mathbf{x}_{i}\) in the scene \(s_i\) and receiving the target description \(q_i\). 
Thus, an episode can be denoted as \(E_i = (s_i, q_i, \mathbf{x}_{i})\). 
At each time step \(t\), the agent has state \(\mathbf{x}^t\), gets the observation \(Z^t\) from the environment, and takes an action. 
At time \(t\), the observation is denoted as
\(Z^t=\{I^t,D^t,\mathbf{T}_t\}\), where \(I^t\) and \(D^t\) are the egocentric
RGB and depth images captured by a limited-FOV RGB-D camera
(\(79^\circ\)), and \(\mathbf{T}_t\) is the odometry-derived pose relative to the starting pose.
The action space consists of six discrete actions, including \texttt{move\_forward}, \texttt{turn\_left}, \texttt{turn\_right}, \texttt{look\_up}, \texttt{look\_down}, and \texttt{stop}.
The \texttt{move\_forward} action moves the agent \(25\,\mathrm{cm}\), while the \texttt{turn\_left}, \texttt{turn\_right}, \texttt{look\_up}, and \texttt{look\_down} actions rotate the agent by \(30^\circ\). 
The \texttt{stop} action is used when the agent believes it has found the target object. 
An episode is considered successful if the agent takes the \texttt{stop} action when it is close enough to the target. 
Each episode has a maximum limit of 500 time steps.

\subsection{The TravExplorer Architecture}

As shown in Fig.~\ref{fig:framework}, TravExplorer is organized as a closed-loop architecture that couples traversability-aware mapping, lightweight semantic guidance, and hierarchical exploration planning. 
At each step, posed RGB-D observations are fused into a volumetric map whose traversability layer is updated from both geometric ray casting and semantic traversability cues. 
The resulting map supports incremental obstacle inflation and frontier extraction directly on connected support surfaces, allowing floors, stairs, and landings to be represented within a unified 3-D planning space. 
In parallel, the semantic guidance module accumulates open-vocabulary detections into a probabilistic instance map and projects image-text relevance onto reachable regions to provide target and stair-related evidence. 
The planning module then reasons over target instances, traversable frontiers, and stair landmarks, selects an admissible exploration goal, searches for a 3-D traversable path, and executes the path with online review and replanning. 
This design keeps high-level semantic decisions grounded in physically executable traversability, which is essential for cross-floor object navigation with a quadruped robot.

\section{3-D Traversable Frontiers}
\begin{figure}[t]
	\centering
	\includegraphics[width=1.0\linewidth]{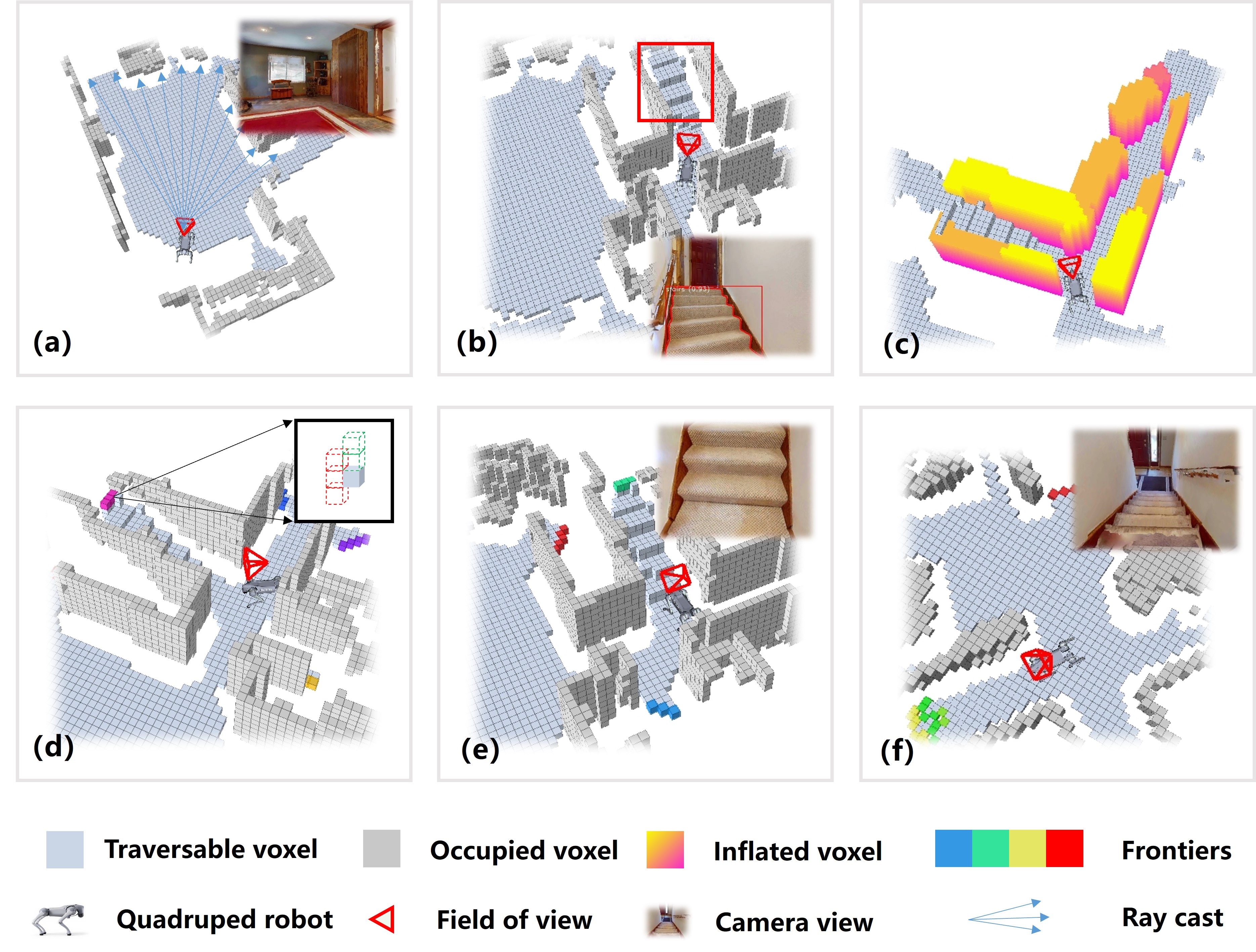}
    \captionsetup{font=footnotesize} 
	\caption{3-D traversable frontier representation.
Blue, gray, and inflated voxels denote traversable support surfaces, occupied structures, and inflated collision regions, respectively, while colored clusters indicate traversable frontiers.
TravExplorer extracts and updates frontiers only on robot-reachable support surfaces, with ray-casting-based updates and FOV-aware active perception to support executable exploration in cluttered and cross-floor environments.
}
	\label{fig:trav_frontier}
\end{figure}

This section introduces the 3-D traversable-frontier representation shown in Fig.~\ref{fig:trav_frontier}, which converts raw RGB-D observations into executable exploration candidates in 3-D traversable space. 
The representation consists of adaptive traversability mapping, incremental obstacle inflation, online frontier updating, and FOV-aware active perception.

\subsection{Adaptive Traversability Mapping}
\label{sec:adaptive_trav_mapping}
TravExplorer maintains a mode-adaptive traversability layer that is efficient for regular planar exploration while remaining complete enough for non-planar floor transitions.
Given an RGB-D observation at time \(t\), denoted by
\(Z^t=\{I^t,D^t,\mathbf{T}_t\}\), where \(\mathbf{T}_t\)
is the camera pose in the world frame, the system maintains a voxel map \(M^t\) that jointly encodes occupancy and traversability.
Each voxel \(v_i\in M^t\) has a center position \(\mathbf{p}_i\)
and an occupancy log-odds value \(l_i^t\). The map maintains two
traversability sources: a geometrically traversable set
\(M_{g}^{t}\) and a semantically traversable set
\(M_{s}^{t}\). The former is updated by efficient 2-D ray
casting on the support plane, while the latter is obtained from
open-vocabulary segmentation and is mainly used for non-planar structures
such as stairs.

The depth image \(D^t\) is first back-projected into a point cloud
\(P^t\) and fused into the occupancy using standard
log-odds updates. 
Voxels with \(l_i^t\ge l_{\mathrm{occ}}\)
are treated as occupied. 
The occupied voxels are further inflated according to the robot body size, forming the inflated obstacle set
\(O_{\mathrm{inf}}^{t}\) to keep a safety margin during traversability update and planning.

Geometric traversability is estimated on the robot support plane. Let \(z_r^t\) denote the height of the current support plane. Points within the robot collision-height range, i.e.,
\(h_{\min}<p_z-z_r^t<h_{\max}\), are projected onto the plane
\(z=z_r^t\). 
For each projected obstacle point, a 2-D ray is cast from the point to the robot support state
\(\mathbf{x}_r^t\). 
Voxels traversed by each ray are marked as geometrically traversable and
inserted into \(M_{g}^{t}\), while occupied voxels are skipped.
This implements 2-D ray casting within the 3-D voxel map.

Semantic traversability is obtained from open-vocabulary segmentation.
Given robot-traversable categories (e.g. floor, stair, and rug), the RGB
image \(I^t\) is segmented to extract traversable regions. 
Corresponding pixels are back-projected using \(D^t\), transformed
into the world frame by \(\mathbf{T}_t\), and the associated voxels are inserted into \(M_{s}^{t}\). 
Open-vocabulary semantic traversability further recovers stairs and ramps that are treated as obstacles by 2-D ray casting, and supports unstructured open-set terrains such as grass.

To balance mapping efficiency and cross-floor completeness, the system
switches between geometric and semantic traversability modes. Let
\(m^t\in\{g,s\}\) denote the mapping mode at time \(t\), where \(g\) and \(s\)
indicate geometric and semantic modes, respectively. 
During regular
exploration, \(m^t=g\), and the planner uses the geometric traversability
induced by 2-D ray casting. 
When stairs are stably detected over
\(N_{\mathrm{enter}}\) consecutive frames, or when the high-level planner
requests navigation to a stair entrance landmark, the system switches to \(m^t=s\),
allowing semantic segmentation to provide traversability over non-planar
support regions.
The final traversable map used by the planner is obtained by selecting the
traversability source consistent with the current mode and removing inflated
obstacles:
\begin{equation}
M_{\mathrm{trav}}^{t}
=
\left(
\mathop{\cup}_{\alpha\in\{g,s\}}
(M_{\alpha}^{t}\mid m^t=\alpha)
\right)
\setminus
O_{\mathrm{inf}}^{t}.
\end{equation}
The system switches back to geometric mode only after the robot remains
sufficiently far from the stair region for \(N_{\mathrm{exit}}\) consecutive
frames. This hysteresis avoids frequent mode oscillations caused by
intermittent stair observations.

\subsection{Incremental Obstacle Inflation}
\label{sec:incremental_obstacle_inflation}

Given the traversability map, the inflation module constructs a planning-safe obstacle layer while avoiding unnecessary inflation of walkable support surfaces.
Uniform 3-D inflation is not suitable for ground robots, since it may incorrectly penalize traversable support regions. 
We therefore inflate only non-traversable occupied voxels. 
In addition, we design an anisotropic inflation neighborhood.
Specifically, the robot is approximated by the center of its bottom support plane. 
The inflation kernel is defined around this reference point,
with a fixed radius in the \(xy\)-plane and a downward-only range along the \(z\)-axis.
After each occupancy and traversability update, the system tracks only voxels whose states have changed. 
The voxels include occupied-to-free transitions and non-traversable-to-traversable transitions:
\begin{equation}
C^{t}
=
\left(O^{t-1}\setminus O^{t}\right)
\cup
\left(M_{\mathrm{trav}}^{t}\setminus M_{\mathrm{trav}}^{t-1}\right).
\end{equation}

For each changed voxel \(v_i\in C^{t}\), previous inflated labels
inside its inflation neighborhood \(N_{\mathrm{inf}}(v_i)\)
are cleared. 
The operation removes obsolete inflated regions caused by newly freed
voxels or newly traversable support regions.
Re-inflation is restricted to the current local update region \(\Omega^t\) and non-traversable voxels. Specifically, the re-inflated region is
\begin{equation}
R_{\mathrm{inf}}^{t}
=
\mathop{\cup}_{v_i\in S_{\mathrm{inf}}^{t}}
N_{\mathrm{inf}}(v_i),
\quad
S_{\mathrm{inf}}^{t}
=
\Omega^t
\cap
O^{t}
\setminus
M_{\mathrm{trav}}^{t}.
\end{equation}

After clearing and re-inflation, the inflated obstacle set
\(O_{\mathrm{inf}}^{t}\) is updated incrementally, while ESDF and frontier updates are then triggered. 
Our incremental inflation strategy has a time complexity of \(O(n)\), where
\(n\) is the number of changed voxels induced by both occupancy and
traversability transitions, avoiding global inflation over the entire 3-D map.

\subsection{Traversable Frontier Updating}
\label{sec:trav_frontier_updating}

The safe traversability map further enables frontier candidates that are both informative for exploration and reachable by the robot.
To support exploration in 3-D traversable spaces, we define frontiers on
robot-reachable support surfaces. Unlike conventional frontiers \cite{zhou2021fuel}, which are
known-free voxels adjacent to unknown space, traversable frontiers describe the
boundary between reachable traversable surfaces and locally unexplored or
temporarily non-traversable regions.

Frontiers are extracted from the traversable map
\(M_{\mathrm{trav}}^t\). A voxel
\(v_i\in M_{\mathrm{trav}}^t\) is regarded as a traversable frontier
candidate if it satisfies two additional conditions.
First, \(v_i\) must have sufficient headroom. No traversable voxel should
appear within \(K_h\) layers above \(v_i\), which avoids extracting frontiers
inside vertical structures or overlapping traversable layers.
Second, \(v_i\) must lie on the boundary of the current traversable surface.
Let \(N_4(v_i)\) denote the horizontal four-neighborhood of
\(v_i\). The voxel \(v_i\) is selected if there exists a neighbor
\(v_j\in N_4(v_i)\) whose vertical column is not traversable. This
places the frontier at the boundary between the reachable support surface and
unexplored or temporary non-traversable space.
The resulting candidates are grouped by 26-neighbor connectivity. 
A traversable frontier cluster is defined as
$
F_i
=
\left\langle
C_i,
\mathbf{p}_i,
B_i
\right\rangle ,
$
where \(C_i=\{v_1,v_2,\ldots,v_{n_i}\}\) denotes the set of
frontier voxels in cluster \(i\), \(\mathbf{p}_i\in\mathbb{R}^3\) is the
geometric centroid of \(C_i\), and \(B_i\) is the
axis-aligned bounding box (AABB) of the cluster. 
The centroid \(\mathbf{p}_i\) is used
as the representative point for goal selection and path-cost evaluation.
Clusters smaller than a predefined size threshold are discarded.

As the traversable map \(M_{\mathrm{trav}}^t\) is updated by new
sensor measurements, the updated region is recorded by its AABB
\(B_m^t\). 
Frontier updating is performed only around this affected
region to support real-time updates, following \cite{zhou2021fuel}. 
The algorithm first traverses all existing frontier clusters and
returns those whose AABBs \(B_i\) intersect with
\(B_m^t\). 
These affected clusters are removed, and their visited
flags are reset. 
New frontier clusters are then extracted in the expanded
neighborhood of \(B_m^t\) by region growing with 26-neighbor connectivity. 
Clusters outside the affected region are kept unchanged, avoiding
global frontier re-extraction over the entire 3-D map.

\subsection{FOV-Aware Active Perception}

Limited forward-looking camera FOV motivates active perception around stair transitions and close-range frontier regions.
Nearby traversable regions may remain unobserved during exploration.
This issue is common when the robot approaches staircases from narrow corridors, where the first stair step may fall outside
the forward-looking view. As a result, nearby frontier clusters remain active even when the robot is already close to the corresponding region.

To handle this issue, we introduce an active perception strategy.
For each active frontier cluster \(F_i\in F_{\mathrm{act}}\), its
distance to the robot is computed as
\(d_i=\|\mathbf{p}_i-\mathbf{x}_r^t\|\), where \(\mathbf{p}_i\) is the
representative point of \(F_i\). If \(d_i<d_{\mathrm{blind}}\), the cluster is
regarded as a blind-spot frontier. The robot then actively pitches the camera
downward to observe the nearby unscanned region and update the traversable map.
If the hidden traversable voxels are observed, the corresponding frontier is
removed by the local frontier update. Otherwise, the frontier is regarded as an
irreducible frontier and moved from the active set
\(F_{\mathrm{act}}\) to the dormant set \(F_{\mathrm{dor}}\)
to avoid repeated ineffective actions.

Descending stairs also require active perception, since they may be completely outside the forward-looking view. 
We introduce a virtual ground plane \(\Pi_{\mathrm{vg}}\) below the robot. 
During exploration, if the number of observed points below \(\Pi_{\mathrm{vg}}\)
exceeds a threshold \(N_{\mathrm{down}}\), the robot triggers a downward-looking
behavior. 
This allows the robot to verify descending staircases and maintain a downward view during stair descent.

\section{Spatial Semantic Guidance}
Semantic guidance provides target awareness on top of traversable-frontier exploration. 
Instead of invoking large-model reasoning at every planning step, TravExplorer maintains two lightweight map-level memories: a probabilistic instance map for persistent object and landmark hypotheses, and a spatial value map that projects image-text relevance onto reachable support surfaces.

\subsection{Probabilistic Instance Map}
Single-frame open-vocabulary detections are accumulated into persistent 3-D semantic hypotheses before being used for planning.
Open-vocabulary perception provides zero-shot recognition ability, but single-frame predictions are often unstable due to false positives, missed detections, occlusions, and viewpoint changes. We maintain a probabilistic semantic instance map. 
The object branch accumulates target-object
hypotheses for goal-directed navigation, while the landmark branch accumulates navigation-related semantic landmarks, such as stairs, for cross-floor guidance.

\begin{figure}[t]
	\centering
	\includegraphics[width=1.0\linewidth]{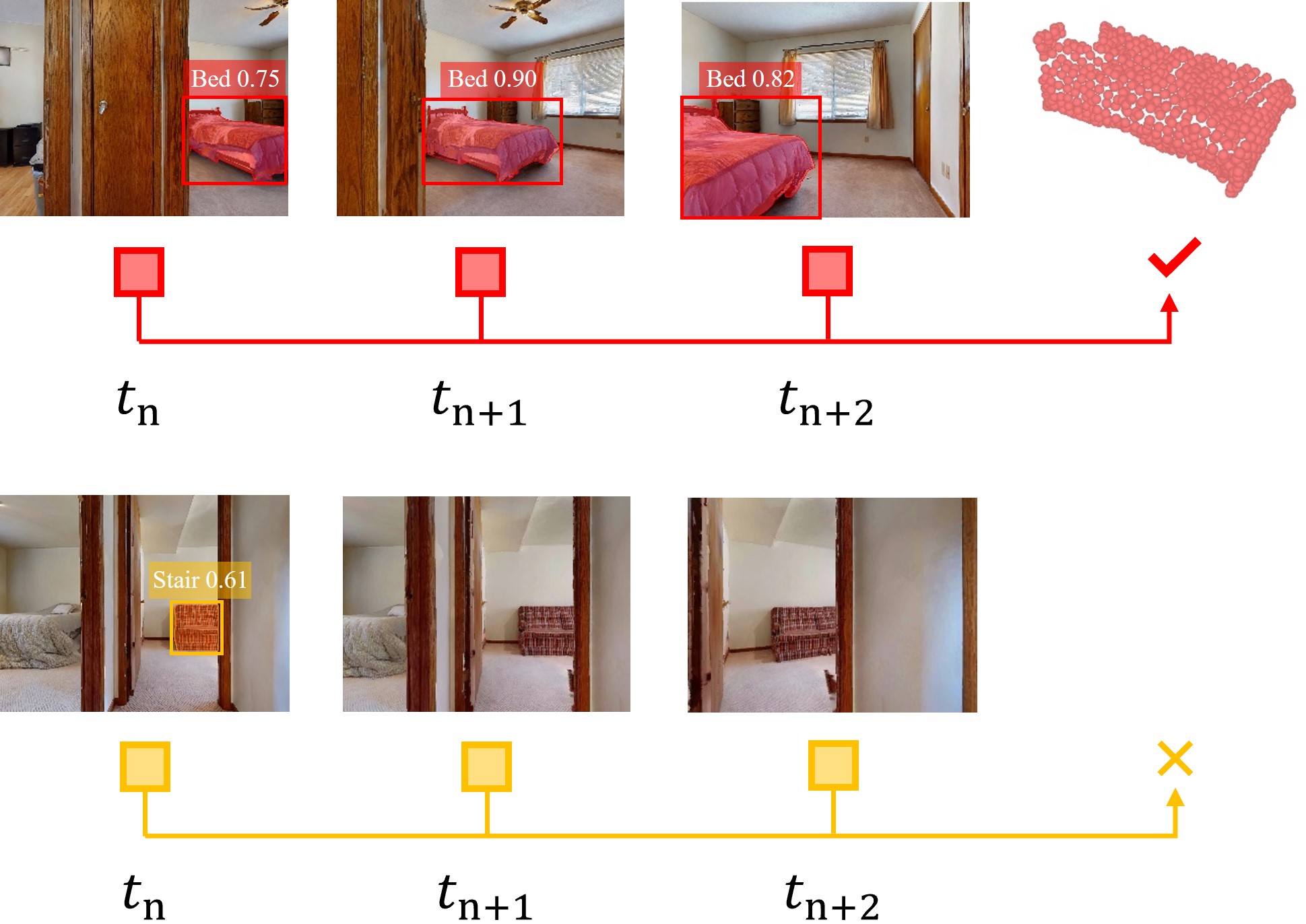}
    \captionsetup{font=footnotesize} 
	\caption{Target-centric instance memory for robust open-vocabulary perception.
TravExplorer accumulates temporally consistent object detections over time and back-projects them into a unified 3-D instance representation.
Persistent detections of the queried object, such as the bed shown in the top row, are fused into a valid target instance, while transient or inconsistent semantic responses, such as the false stair detection in the bottom row, are rejected.
This temporal instance memory reduces single-view false positives and provides more reliable semantic evidence for target-guided exploration.
}
	\label{fig:instance_map}
\end{figure}

The query set is defined as
\(Q=Q_{\mathrm{obj}}\cup Q_{\mathrm{land}}\),
where \(Q_{\mathrm{obj}}\) contains target-object queries and
\(Q_{\mathrm{land}}\) contains traversability-related landmark
queries. At time \(t\), for each query \(q\in Q\), the
open-vocabulary segmentation module extracts an image region
\(R_q^t\) from \(I^t\). Pixels in \(R_q^t\) are
back-projected using \(D^t\) to obtain a semantic point cloud
\(P_q^t\). Each detection is represented as
\(d_q^t=(P_q^t,s_q^t,q)\), where \(s_q^t\) is the detection
confidence.

The instance map contains semantic clusters \(Y^t\). Each cluster
\(y\in Y^t\) stores a semantic label \(q_y\), a fused point cloud
\(P_y^t\), a fused confidence \(\gamma_y^t\), and an observed point count
\(n_y^t\). A detection is associated only with clusters of the same label, and
the association is accepted when their 3-D centroid distance is below
\(\delta_{\mathrm{asso}}\).
For an associated cluster, the current detection is treated as positive
evidence. 
Following \cite{zhang2025apexnav}, the point cloud is integrated with voxel downsampling
\(D(\cdot)\), and the confidence is updated by weighted fusion:
\begin{equation}
P_y^t
=
D
\left(
P_y^{t-1}
\cup
P_q^t
\right),
\qquad
n_y^t
=
n_y^{t-1}
+
n_q^t,
\end{equation}
\begin{equation}
\gamma_y^t
=
\frac{
n_y^{t-1}\gamma_y^{t-1}
+
n_q^t s_q^t
}{
n_y^t
},
\qquad
n_q^t=|P_q^t|.
\end{equation}
If no valid association is found, a new cluster is initialized from the current
detection.
To suppress stale false positives, visible but unmatched clusters are updated
with negative evidence. If a cluster \(y\) lies inside the current field of
view but receives no associated detection, its visible support
\(\tilde n_y^t\) is added as zero-confidence evidence:
\begin{equation}
n_y^t
=
n_y^{t-1}
+
\tilde n_y^t,
\qquad
\gamma_y^t
=
\frac{
n_y^{t-1}\gamma_y^{t-1}
}{
n_y^t
}.
\end{equation}
This decreases the confidence of hypotheses that are repeatedly observed but
not consistently detected, while preserving instances supported by stable
multi-frame observations.

The two branches provide different candidates to the hierarchical planner.
Clusters from \(Q_{\mathrm{obj}}\) are inserted into
\(O_{\mathrm{tar}}\), while clusters from
\(Q_{\mathrm{land}}\) are inserted into
\(O_{\mathrm{stair}}\). For each query \(q\), reliable candidates
are selected according to the accumulated belief \(\gamma_y^t n_y^t\). A
cluster is used by the planner only when its fused confidence exceeds a
predefined threshold; otherwise, it is retained as a tentative hypothesis for
later verification.

\begin{figure}[t]
	\centering
	\includegraphics[width=1.0\linewidth]{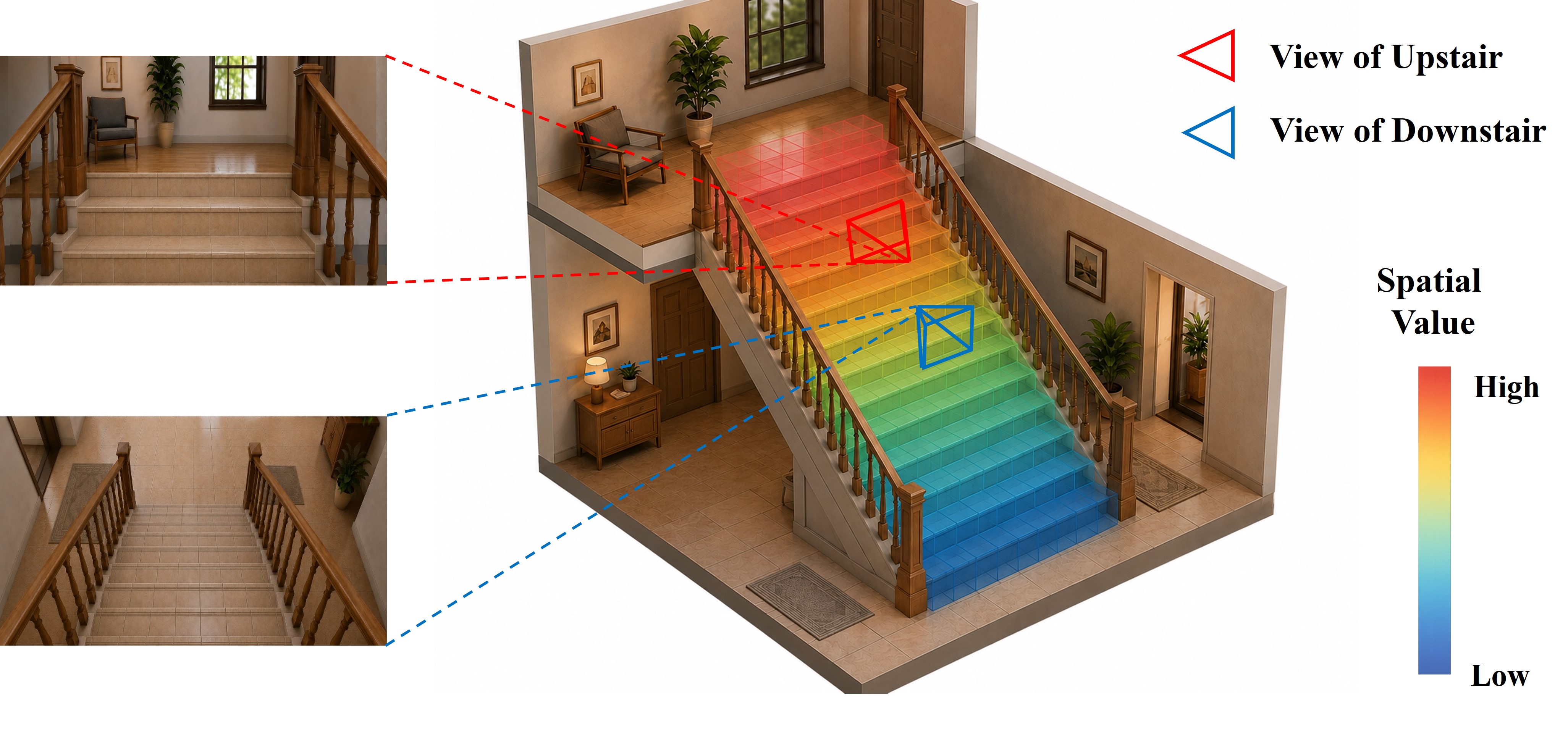}
    \captionsetup{font=footnotesize} 
	\caption{Traversability-aware spatial value map on a staircase.
Image-text relevance from upstairs and downstairs views is projected onto reachable stair surfaces and fused into a 3-D value map.
Higher values indicate regions with stronger target-related evidence, enabling direction-aware semantic guidance for cross-floor exploration.
}
	\label{fig:spatial_value}
\end{figure}

\subsection{Spatial Value Map}
\label{sec:spatial_value_map}

The spatial value map assigns language-conditioned utility to reachable traversable voxels so that frontier selection can exploit semantic evidence before a reliable target instance is confirmed.
Frequent VLM or LLM reasoning in the control loop introduces substantial latency.
Following VLFM~\cite{yokoyama2024vlfm}, we use BLIP-2 \cite{li2023blip2} as a fast image-text matching model for semantic guidance. 
To guide target-driven exploration in
3-D traversable space, we maintain a spatial value map \(V^t=(s^t,r^t)\) over
robot-reachable support regions, where \(s_k^t\in[0,1]\) is the accumulated
spatial semantic value of voxel \(v_k\), and \(r_k^t\in[0,1]\) is its accumulated reliability. 
This map integrates semantic evidence in 3-D and mitigates the
limited spatial scene understanding of VLMs.

At time \(t\), BLIP-2 produces an image-text relevance score \(\hat{s}^t\) for
the current RGB observation. This score is projected only onto visible
traversable voxels. For each visible voxel \(v_k\), the observation weight
\(\omega_k^t=\cos^2(\pi\theta_k^t/\theta_{\mathrm{FOV}})\) is computed from
its horizontal angular offset \(\theta_k^t\) to the camera heading, assigning
larger weights to voxels near the image center and smaller weights near the
FOV boundary. The semantic value and reliability are updated by
confidence-weighted fusion:
\begin{equation}
s_k^t
=
\frac{
r_k^{t-1}s_k^{t-1}
+
\omega_k^t \hat{s}^t
}{
r_k^{t-1}
+
\omega_k^t
},
\qquad
r_k^t
=
\frac{
(r_k^{t-1})^2
+
(\omega_k^t)^2
}{
r_k^{t-1}
+
\omega_k^t
}.
\end{equation}

This update accumulates language-conditioned evidence from multiple viewpoints
and suppresses unstable single-frame predictions. As illustrated in
Fig.~\ref{fig:spatial_value}, semantic evidence from upstairs and downstairs
viewpoints is fused on the corresponding reachable stair surfaces, forming a
coherent spatial belief for downstream planning. For a candidate node, its
semantic utility is computed from the maximum value in its local traversable
neighborhood. A higher value indicates that the candidate frontier is more
likely to guide the robot toward the target object.

\begin{figure*}[t]
		\centering
	    \includegraphics[width=0.95\textwidth]{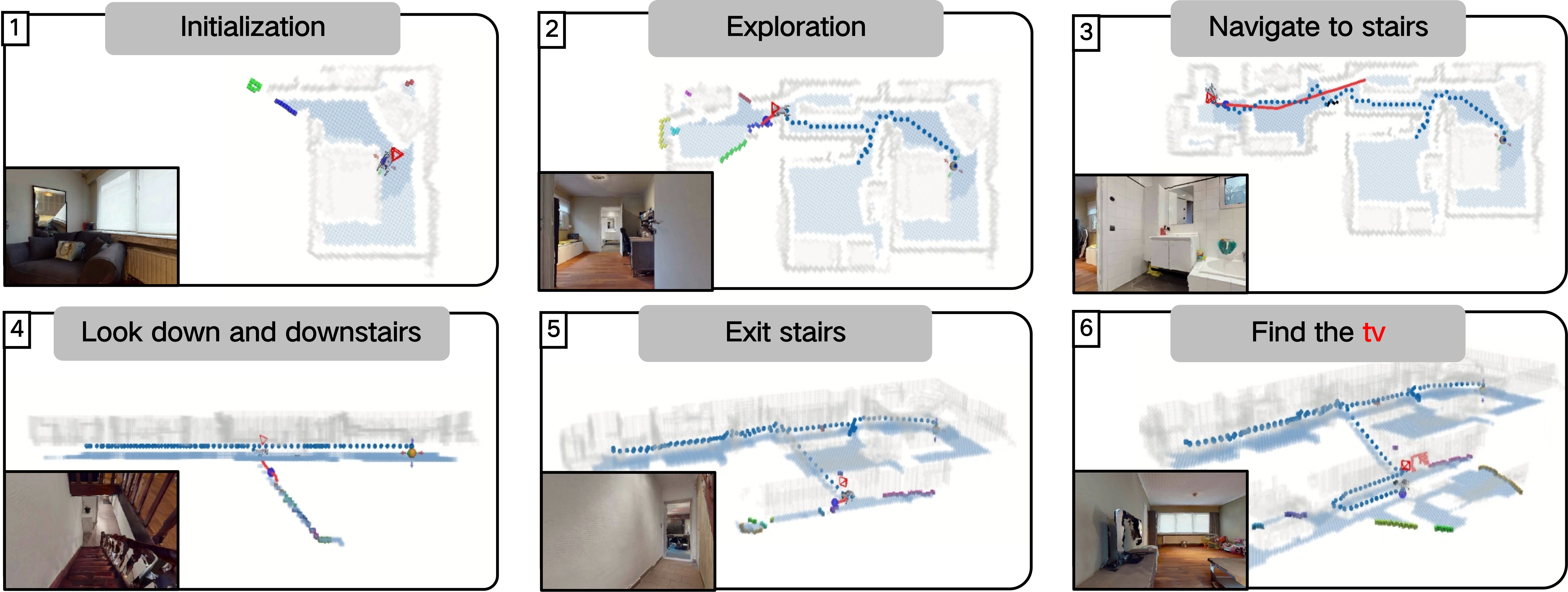}
	    \captionsetup{font=footnotesize}
		\caption{Cross-floor exploration process.
(a) The robot initializes the map and frontier candidates from early RGB-D observations.
(b) TravExplorer expands the known region by following reachable traversable frontiers.
(c) The planner detects a stair-related semantic landmark and navigates toward the stair region.
(d) The robot actively looks downward to complete FOV-limited stair observations and update traversability.
(e) After crossing the stairs, the system preserves the unified 3-D map and continues exploration on the new floor.
(f) Guided by open-vocabulary semantic evidence, the robot reaches the queried target, \textit{tv}.
}
\label{fig:sim_sequence}
\end{figure*}
\section{Hierarchical Exploration Planning}
Given the traversable map and semantic memories, the planner must decide both which exploration candidate to visit and how to reach it safely. 
TravExplorer addresses this with a hierarchical pipeline: an open-tour selects among target instances, traversable frontiers, and stair landmarks; a 3-D graph search provides a route over traversable voxels; an execute-review loop monitors progress; and a local trajectory optimizer generates safe commands for real-world deployment.

\subsection{Chain of Exploration}
\label{sec:chain_of_exploration}

Goal selection follows a hierarchical open-ATSP formulation. At each replanning
step, the robot first maintains a unified candidate pool that includes the
current robot state, semantic instances, and frontier clusters:
\begin{equation}
X
=
\{\mathbf{x}_r\}
\cup
O
\cup
F .
\end{equation}
Here, \(\mathbf{x}_r\) is the current robot state. The semantic candidate set is
\(O=O_{\mathrm{tar}}\cup O_{\mathrm{stair}}\), where \(O_{\mathrm{tar}}\)
contains reliable target-object instances and \(O_{\mathrm{stair}}\) contains
stair-entry landmarks. The frontier candidate set is
\(F=F_g\cup F_s\), where \(F_g\) and \(F_s\) denote geometric and semantic
frontiers on the current floor, respectively.

Instead of optimizing over all candidates simultaneously, the planner activates
one admissible subset according to a lexicographic exploration priority:
\begin{equation}
A
=
\begin{cases}
O_{\mathrm{tar}},
& O_{\mathrm{tar}}\neq\emptyset,\\
F,
& O_{\mathrm{tar}}=\emptyset,\ F\neq\emptyset,\\
O_{\mathrm{stair}},
& O_{\mathrm{tar}}=\emptyset,\ F=\emptyset .
\end{cases}
\end{equation}
The actual ATSP node set is then
\begin{equation}
X_A = \{\mathbf{x}_r\}\cup A .
\end{equation}
This formulation decouples candidate maintenance from goal optimization:
navigation cues are maintained in a unified pool, whereas the open ATSP is
solved only over the subset activated by the current exploration stage.

Navigation is formulated as an open Asymmetric Traveling Salesperson Problem
(ATSP) \cite{meng2017two} over the active node set \(X_A\). Let \(A\) denote
the active candidate nodes. The cost matrix is constructed as
\begin{equation}
C_G
=
\begin{bmatrix}
c_{r,r} & c_{r,A}^{\top} \\
c_{A,r} & C_{A,A}
\end{bmatrix},
\end{equation}
where \(c_{r,r}=0\) and \(c_{A,r}=\mathbf{0}\). This removes the cost of
returning to the current robot state and yields an open tour.

For two active candidate nodes \(v_i\) and \(v_j\), the candidate-to-candidate
cost is
\begin{equation}
C_{A,A}(i,j)
=
\begin{cases}
0, & i=j,\\
J_{\mathrm{geo}}(v_i,v_j)
+
J_{\mathrm{front}}(v_j),
& i\neq j .
\end{cases}
\end{equation}
The cost from the current robot state to an active candidate is
\begin{equation}
c_{r,A}(j)
=
J_{\mathrm{geo}}(\mathbf{x}_r,v_j)
+
J_{\mathrm{front}}(v_j).
\end{equation}
Here, \(J_{\mathrm{geo}}\) is the shortest traversable path cost on the current
traversability map. If no feasible path exists, \(J_{\mathrm{geo}}\) is set to
\(M_\infty\).
The frontier preference term is defined as
\begin{equation}
J_{\mathrm{front}}(v)
=
\begin{cases}
\lambda_g, & v\in F_g,\\
\lambda_s, & v\in F_s,\\
0, & v\notin F,
\end{cases}
\qquad
0\leq \lambda_g < \lambda_s .
\end{equation}
This term provides a soft intra-frontier preference: when the active set is the
current-floor frontier set \(F\), geometric frontiers are preferred over
semantic frontiers, while semantic frontiers serve as complementary candidates
for non-planar traversable regions that may otherwise cause redundant revisits.

The hierarchical activation implements the exploration priority before ATSP optimization. 
Target objects dominate all other navigation cues. 
If no reliable target is available, the planner selects from the current-floor frontier set \(F\). 
Within \(F\), \(J_{\mathrm{front}}\) softly prefers geometric frontiers over semantic frontiers. 
Stair landmarks are activated only after current-floor frontiers are exhausted.
They trigger a cross-floor transition stage, during which semantic frontiers
guide the robot through non-planar stair regions where geometric frontiers are
not continuously defined.
Once the robot reaches a new floor, the frontier candidate pool is updated and the same decision
process is repeated, yielding the exploration chain.

\subsection{3-D Path Search}
\label{sec:3d_path_search}

After a navigation goal is selected, the planner computes a physically executable route on the traversable support graph rather than in the full 3-D free-space grid.
The search is performed on the traversable map \(M_{\mathrm{trav}}^t\), where inflated obstacles have been removed. 
Compared with 26-neighbor expansion in dense 3-D voxel space, expansion over traversable voxels is more efficient.
Due to the limited camera FOV, traversable voxels on vertical transition structures may be locally disconnected. 
Foothold-scale neighborhoods are employed to bridge such small gaps and preserve connectivity across stairs and floors.

For two connected traversable voxels \(v_i\) and \(v_j\), the transition cost is
defined as
\begin{equation}
J_{\mathrm{step}}(v_i,v_j)
=
r\|\Delta \mathbf{k}_{ij}\|_2
+
\lambda_{\mathrm{esdf}}\Phi(v_j),
\end{equation}
where \(r\) is the voxel resolution, \(\Delta\mathbf{k}_{ij}\) is the
voxel-index offset from \(v_i\) to \(v_j\), and
\(\lambda_{\mathrm{esdf}}\) is the ESDF safety weight. The safety penalty is
\begin{equation}
\Phi(v_j)
=
\max
\left(
0,
\frac{d_{\mathrm{safe}}-d_{\mathrm{obs}}(v_j)}
{d_{\mathrm{safe}}}
\right),
\end{equation}
where \(d_{\mathrm{obs}}(v_j)\) is the ESDF distance to the nearest obstacle
and \(d_{\mathrm{safe}}\) is the safety distance. 
This cost favors short transitions with sufficient obstacle clearance.

For ATSP construction, geometric costs are obtained by running a multi-target
graph search from each source candidate to all other candidates. The returned
path lengths directly fill \(J_{\mathrm{geo}}(\cdot,\cdot)\) in the ATSP
matrix, avoiding repeated pairwise searches. After the next navigation goal is
selected, the corresponding 3-D path is recovered by backtracking the parent
nodes from the selected target.

\subsection{Execute-Review Mechanism}
\label{sec:execute_review}

Execution is treated as a closed-loop validation stage that detects disagreement between the planned motion and the observed robot motion.
After a 3-D path is obtained, the system converts path following into
executable local actions. In simulation with a discrete action space, the robot
selects a local waypoint on the planned path and chooses the next action
according to the heading difference between its current yaw and the direction
toward this waypoint. In this way, the robot follows the 3-D path through
discrete motion primitives rather than directly executing a continuous
trajectory.

To monitor execution consistency, the system maintains a predicted pose
\(\hat{\mathbf{x}}_r^t\) from the action model. After an action is issued,
\(\hat{\mathbf{x}}_r^t\) is updated according to the corresponding motion
primitive. The executed motion is then checked against the odometry pose
\(\mathbf{x}_{\mathrm{odom}}^t\). If a forward action satisfies
\begin{equation}
\left\|
\mathbf{x}_{\mathrm{odom}}^t
-
\hat{\mathbf{x}}_r^t
\right\|_2
>
\delta_{\mathrm{stuck}},
\end{equation}
the robot is considered blocked by an unmodeled obstacle or a local mapping
error.
When such a failure occurs, the voxel in front of the robot, denoted by
\(v_{\mathrm{front}}\), is conservatively marked as occupied. The traversable
map is then updated, and the planner replans on the new map. If similar
failures repeatedly occur near the same target, the corresponding frontier is
moved to the dormant set to avoid repeatedly selecting unreachable regions.

Notably, for real-world settings where discrete actions are no longer applicable, this execution layer is replaced by local trajectory optimization to generate safe and continuous spatiotemporal commands.

\subsection{Local Trajectory Optimization}
\label{sec:local_traj_opt}

For real-world deployment, the discrete traversable path must be converted into a smooth and dynamically feasible command sequence while preserving the 3-D support-surface profile.
The searched 3-D path is refined by a local
B-spline trajectory optimizer. The graph-search path is collision-free but
piecewise linear due to voxel discretization. The optimizer smooths a local
path segment while preserving the ground-following structure encoded by the
traversability-aware path.

Let the searched traversable path be
\(\Gamma=\{\mathbf{x}_0,\mathbf{x}_1,\ldots,\mathbf{x}_M\}\), where
\(\mathbf{x}_i=[x_i,y_i,z_i]^\top\in\mathbb{R}^3\). At each replanning step, a
local target \(\mathbf{x}_g=[x_g,y_g,z_g]^\top\) is selected from \(\Gamma\)
within a look-ahead distance from the current robot state
\(\mathbf{x}_s=[x_s,y_s,z_s]^\top\).

A local initial segment is generated between \(\mathbf{x}_s\) and
\(\mathbf{x}_g\). The horizontal components are generated by polynomial
interpolation according to the
current motion state and the selected target, while the height is interpolated
along the accumulated horizontal arc length:
\begin{equation}
z_i
=
z_s
+
\frac{\sigma_i}{\sigma_K}
\left(z_g-z_s\right),
\end{equation}
where \(\sigma_i\) is the accumulated horizontal arc length from
\(\mathbf{x}_s\) to the \(i\)-th sample, and \(\sigma_K\) is the total
horizontal arc length of the local segment.
The initialized samples are converted into cubic B-spline control points
\(\{\mathbf{Q}_0,\mathbf{Q}_1,\ldots,\mathbf{Q}_N\}\). Given the knot span
\(\Delta t\), the corresponding velocity and acceleration control points are
\(\mathbf{v}_i=(\mathbf{Q}_{i+1}-\mathbf{Q}_i)/\Delta t\) and
\(\mathbf{a}_i=(\mathbf{v}_{i+1}-\mathbf{v}_i)/\Delta t\), respectively.
Following \cite{zhou2020ego}, the trajectory is optimized by minimizing
\begin{equation}
J_{\mathrm{total}}
=
\lambda_s J_s
+
\lambda_o J_o
+
\lambda_f J_f,
\end{equation}
where
\begin{equation}
J_s
=
\sum_{i=1}^{N-2}
\left\|
\frac{\mathbf{a}_{i+1}-\mathbf{a}_i}{\Delta t}
\right\|^2,
\end{equation}
\begin{equation}
J_o
=
\sum_{i=0}^{N}
F_{\mathrm{obs}}
\left(
d(\mathbf{Q}_i)
\right),
\end{equation}
and
\begin{equation}
J_f
=
\sum_{i=0}^{N-1}
F_{\mathrm{lim}}
\left(
\|\mathbf{v}_i\|,
v_{\max}
\right)
+
\sum_{i=1}^{N-1}
F_{\mathrm{lim}}
\left(
\|\mathbf{a}_i\|,
a_{\max}
\right).
\end{equation}
Here, \(J_s\) penalizes jerk-like variation of acceleration control points,
\(J_o\) penalizes proximity to obstacles, and \(J_f\) enforces dynamic
feasibility. The obstacle and limit penalties are defined as
\(F_{\mathrm{obs}}(d)=\max(d_{\min}^2-d^2,0)\) and
\(F_{\mathrm{lim}}(x,x_{\max})=\max(x^2-x_{\max}^2,0)\), respectively.

Unlike unconstrained 3-D trajectory optimization, the trajectory is not freely
deformed away from the traversable surface. Since \(\Gamma\) already
provides a ground-following reference over floors and stairs,, only
the horizontal components of the B-spline control points are optimized. For
each control point \(\mathbf{Q}_i\), the objective gradient is projected onto
the horizontal subspace:
\begin{equation}
\tilde{\nabla}_{\mathbf{Q}_i}J_{\mathrm{total}}
=
\mathbf{P}_{xy}
\nabla_{\mathbf{Q}_i}J_{\mathrm{total}},
\qquad
\mathbf{P}_{xy}
=
\mathrm{diag}(1,1,0).
\end{equation}
This projected update improves horizontal smoothness and clearance while preserving the height profile induced by the 3-D traversability-aware path.

\section{Results}
\subsection{Experimental Setup}

We evaluate TravExplorer in both simulation benchmarks and real-world environments. 
In simulation, we conduct 4195 episodes across two benchmarks.
In the real world, we conduct 50 episodes, including 32 episodes for single-floor object navigation and 18 episodes for cross-floor object navigation.
Notably, extensive benchmark comparisons and real-world experiments are demonstrated in the \textbf{supplementary videos}.

\textbf{Simulation benchmark setup.}
In simulation, we follow the Habitat Challenge 2023 setup~\cite{habitat_challenge_2023}. 
The robot is equipped with a $640 \times 480$ RGB-D camera, mounted at a height of $0.88\,\mathrm{m}$.
The agent moves forward by $0.25\,\mathrm{m}$ per step and rotates by $30^\circ$ per action. 
The sensor range is $[0.0\,\mathrm{m}, 5.0\,\mathrm{m}]$, and the success distance is $0.2\,\mathrm{m}$. 
We conduct experiments on two benchmarks: HM3D~\cite{ramakrishnan2021hm3d}, with 2000 episodes across 20 high-fidelity scenes and 6 target categories; MP3D~\cite{chang2017matterport3d}, with 2195 episodes across 11 scenes and 21 target categories.

\begin{table}[h]
    \centering
    \renewcommand{\arraystretch}{1.1} 
    \setlength{\tabcolsep}{3.5pt}     
    \caption{Comparisons with state-of-the-art methods. The table contrasts learning-based and zero-shot methods on HM3D and MP3D datasets.}
    \label{tab:sota}
    
    \begin{tabular}{c @{\hspace{6pt}} c @{\hspace{6pt}} c c c c c}
        \toprule
        \multirow{2}{*}{\textbf{Method}} & \multirow{2}{*}{\textbf{Zero-shot}} & \multirow{2}{*}{\textbf{Cross-Floor}} & \multicolumn{2}{c}{\textbf{HM3D}} & \multicolumn{2}{c}{\textbf{MP3D}} \\
        \cmidrule(lr){4-5} \cmidrule(lr){6-7}
        & & & SR$\uparrow$ & SPL$\uparrow$ & SR$\uparrow$ & SPL$\uparrow$ \\
        \midrule
        SemExp~\cite{chaplot2020object}      & $\times$ & $\times$ & -- & -- & 36.0 & 14.4 \\
        PONI~\cite{ramakrishnan2022poni}       & $\times$ & $\times$ & -- & -- & 31.8 & 12.1 \\
        ZSON~\cite{majumdar2022zson}       & $\times$ & $\times$ & 25.5 & 12.6 & 15.3 & 4.8 \\
        \midrule
        CoW~\cite{gadre2023cows}         & $\checkmark$ & $\times$ & -- & -- & 7.4  & 3.7 \\
        ESC~\cite{zhou2023esc}        & $\checkmark$ & $\times$ & 39.2 & 22.3 & 28.7 & 14.2 \\
        L3MVN~\cite{yu2023l3mvn}      & $\checkmark$ & $\times$ & 50.4 & 23.1 & 34.9 & 14.5 \\
        OpenFMNav~\cite{kuang2024openfmnav}   & $\checkmark$ & $\times$ & 54.9 & 24.4 & 37.2 & 15.7 \\
        VLFM~\cite{yokoyama2024vlfm}       & $\checkmark$ & $\times$ & 52.5 & 30.4 & 36.4 & 17.5 \\
        TriHelper~\cite{zhang2024trihelper}  & $\checkmark$ & $\times$ & 56.5 & 25.3 & -- & -- \\
        SG-Nav~\cite{yin2024sgnav}     & $\checkmark$ & $\times$ & 54.0 & 24.9 & 40.2 & 16.0 \\
        ApexNav~\cite{zhang2025apexnav}    & $\checkmark$ & $\times$ & 59.6 & 33.0 & 39.2 & 17.8 \\
        \midrule
        MFNP~\cite{zhang2024mfnp}       & $\checkmark$ & $\checkmark$ & 58.3 & 26.7 & 41.1 & 15.4 \\
        ASCENT~\cite{gong2026ascent}     & $\checkmark$ & $\checkmark$ & 65.4 & 33.5 & 44.5 & 15.5 \\
        \rowcolor{gray!15} \textbf{TravExplorer} & $\checkmark$ & $\checkmark$ & \textbf{70.0} & \textbf{37.2} & \textbf{48.8} & \textbf{21.1} \\
        \bottomrule
    \end{tabular}
\end{table}

\textbf{Evaluation metrics.}
Following~\cite{habitat_challenge_2023}, we use two primary metrics: Success Rate (SR), defined as the percentage of episodes in which the robot reaches the target object within a predefined success distance, and Success weighted by Path Length (SPL), which penalizes successful episodes by the ratio between the shortest path length and the actual path length. 
For real-world experiments, where ground-truth shortest path lengths are unavailable, we additionally report Success Penalized by Time (SPT),
\begin{equation}
    \mathrm{SPT}
    =
    \mathrm{SR}
    \left(
    1 - \frac{t}{T_{\mathrm{timeout}}}
    \right),
\end{equation}
where $t$ denotes the task completion time and $T_{\mathrm{timeout}}$ denotes the predefined timeout threshold. We also report the Average Time (AT) over successful episodes.

\textbf{Implementation details.}
For semantic segmentation, we use SAM~3~\cite{carion2025sam3}. 
The semantic score map is generated by BLIP-2~\cite{li2023blip2}.
The ATSP is solved using the LKH solver \cite{helsgaun2017extension}.
All methods are evaluated under the same benchmark protocol, including identical episode splits, sensor configuration, success criterion, and target categories for fair comparison.
All simulation experiments are run on a workstation with an NVIDIA GeForce RTX 4090 GPU and an Intel i9-14900K $\times 32$ CPU.

\subsection{Quantitative Results}
\textbf{Baselines.}
We compare TravExplorer with representative learning-based, zero-shot, and floor-aware ObjectNav baselines. SemExp~\cite{chaplot2020object}, PONI~\cite{ramakrishnan2022poni}, and ZSON~\cite{majumdar2022zson} use learned policies or navigation priors trained with task-specific supervision, which limits zero-shot transfer. CoW~\cite{gadre2023cows} performs nearest-frontier exploration without explicit semantic guidance. ESC~\cite{zhou2023esc}, L3MVN~\cite{yu2023l3mvn}, TriHelper~\cite{zhang2024trihelper}, and OpenFMNav~\cite{kuang2024openfmnav} incorporate semantic maps and LLM-based frontier selection. VLFM~\cite{yokoyama2024vlfm} and InstructNav~\cite{long2024instructnav} construct vision-language value maps for target-directed exploration. SG-Nav~\cite{yin2024sgnav} uses 3-D scene-graph prompting to improve semantic frontier reasoning. ApexNav~\cite{zhang2025apexnav} adds target-centric semantic fusion for adaptive 2-D exploration, whereas MFNP~\cite{zhang2024mfnp} and ASCENT~\cite{gong2026ascent} address multi-floor ObjectNav through floor-aware exploration policies.

\textbf{Comparisons with SOTAs.}
Table~\ref{tab:sota} reports the comparison with representative learning-based and zero-shot ObjectNav methods on HM3D and MP3D. 
TravExplorer achieves the best performance on both datasets, with $70.0\%$ SR and $37.2\%$ SPL on HM3D, and $48.8\%$ SR and $21.1\%$ SPL on MP3D. 
Compared with ASCENT~\cite{gong2026ascent}, the strongest cross-floor baseline, TravExplorer improves SR/SPL by $+4.6\%/+3.7\%$ on HM3D and $+4.3\%/+5.6\%$ on MP3D. 
Compared with ApexNav~\cite{zhang2025apexnav}, a strong zero-shot baseline based on single-floor 2-D exploration, TravExplorer further improves SR/SPL by $+10.4\%/+4.2\%$ on HM3D and $+9.6\%/+3.3\%$ on MP3D.

The results show that the proposed method improves both task completion and path efficiency. 
In particular, most zero-shot baselines, such as VLFM and ApexNav, construct exploration policies on top of 2-D frontier maps, which are effective for planar navigation but cannot explicitly represent vertical connectivity. 
MFNP and ASCENT introduce floor-aware mechanisms to handle multi-floor environments, yet their representations remain essentially layer-wise and depend on discrete floor transitions. 
In contrast, TravExplorer represents the environment as a unified 3-D traversable space, where floors, stairs, and intermediate landings are connected through an explicit traversability structure. 
This representation avoids the ambiguity caused by projecting multi-floor geometry onto 2-D maps and enables
the planner to reason directly about cross-floor traversability.
The consistent gains on both HM3D and MP3D indicate that explicit 3-D traversability reasoning is beneficial not only for cross-floor reachability, but also for more efficient target-directed exploration.

\begin{figure}[t]
	\centering
	\includegraphics[width=1.0\linewidth]{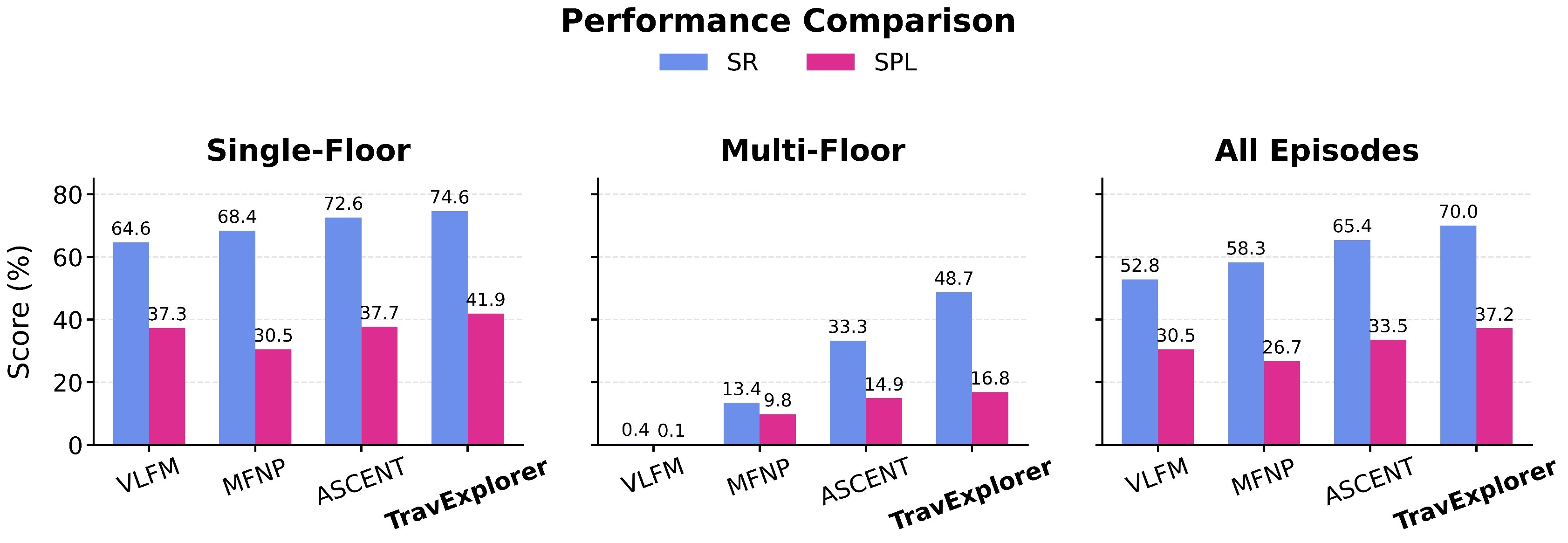}
    \captionsetup{font=footnotesize} 
	\caption{Performance comparison on HM3D across single-floor, multi-floor, and all-episode settings.
    We report Success Rate (SR) and Success weighted by Path Length (SPL) for representative zero-shot and floor-aware ObjectNav baselines.
    TravExplorer achieves the best overall performance with $70.0\%$ SR and $37.2\%$ SPL, while maintaining strong single-floor performance.
    The improvement is most pronounced in multi-floor episodes, where the unified 3-D traversability representation and cross-floor planning enable more reliable reasoning over stairs, landings, and vertically connected spaces.
}
	\label{fig:multi_ablation}
\end{figure}

\textbf{Performance Analysis Across Floor Scenarios.}
Fig.~\ref{fig:multi_ablation} further analyzes performance on HM3D by separating single-floor and multi-floor episodes. 
In single-floor episodes, TravExplorer achieves $74.6\%$ SR and $41.9\%$ SPL, outperforming ASCENT by $+2.0\%$ SR and $+4.2\%$ SPL. 
This suggests that introducing a 3-D traversability representation does not degrade conventional planar ObjectNav performance. 
Instead, the target-centric semantic fusion and value-guided frontier touring help reduce premature target commitments and redundant exploration, leading to higher path efficiency.

The advantage becomes more pronounced in multi-floor episodes. 
TravExplorer achieves $48.7\%$ SR and $16.8\%$ SPL, exceeding VLFM, MFNP, and ASCENT by $+48.3\%$, $+35.3\%$, and $+15.4\%$ SR, respectively. 
The comparison also explains the limitations of existing representations. 
VLFM does not explicitly model floor transitions, making it unsuitable for cross-floor navigation. 
MFNP relies on floor switching with map reset, which can break spatial continuity and lose useful exploration history. 
ASCENT improves cross-floor reasoning but still operates on floor-wise 2-D abstractions, where stairs and landings are not represented as continuous traversable structures. 
By combining active stair perception with foothold-guided 3-D search, TravExplorer maintains continuous traversability across floors and therefore achieves more reliable cross-floor exploration while retaining competitive SPL.

\begin{figure}[t]
	\centering
	\includegraphics[width=1.0\linewidth]{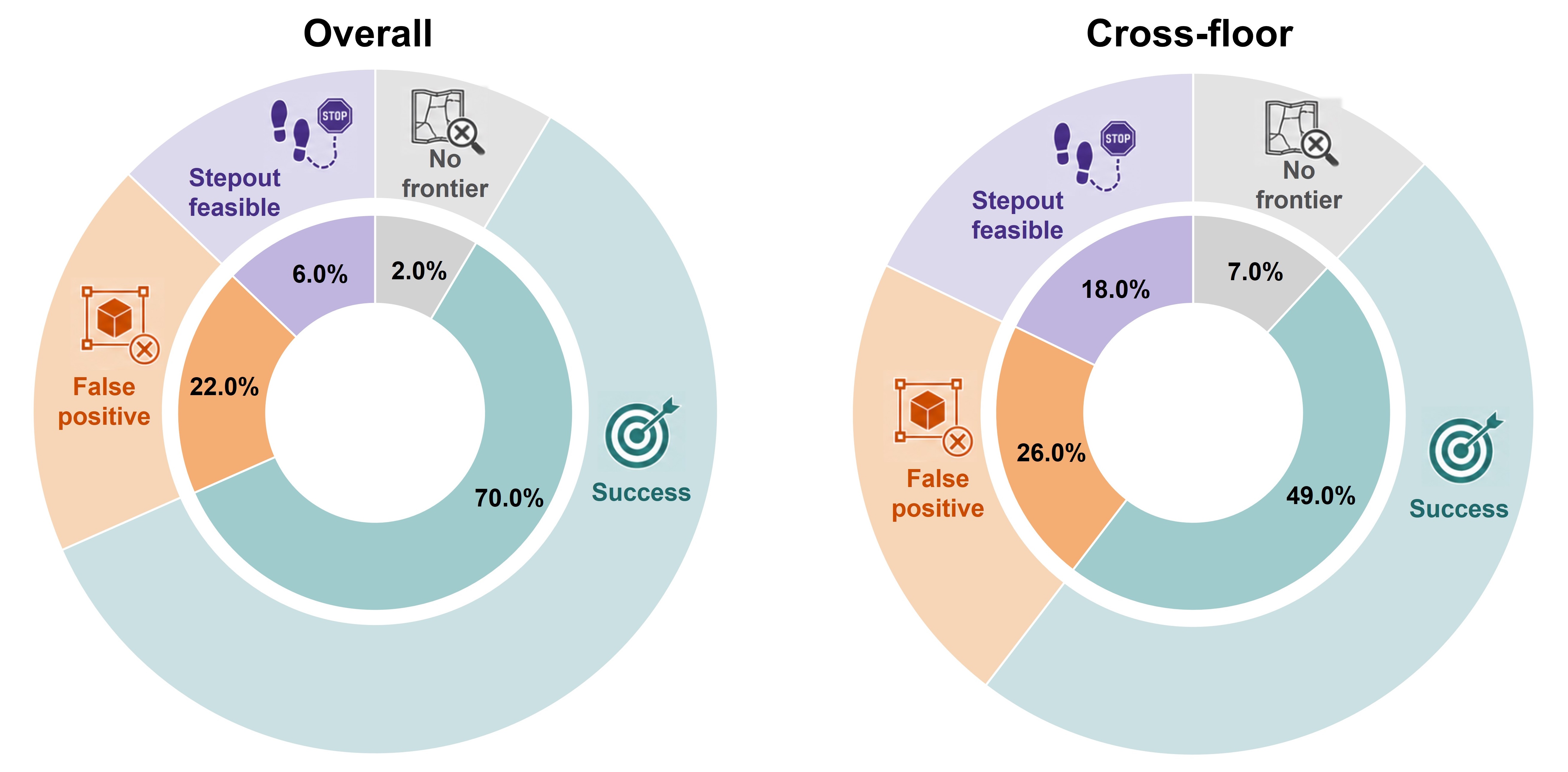}
    \captionsetup{font=footnotesize} 
	\caption{Failure-mode analysis on HM3D.
TravExplorer achieves a $70.0\%$ success rate over all episodes and a $49.0\%$ success rate on cross-floor episodes.
The remaining failures are mainly caused by false-positive semantic detections, step-out feasibility violations, and missing frontiers, with cross-floor scenes introducing more feasibility-related challenges due to stairs, landings, and incomplete 3-D observations.
}
	\label{fig:failure_modes}
\end{figure}

\textbf{Failure Case Analysis.}
Outcomes are categorized as follows:
\begin{itemize}
    \item \textbf{Success}: The robot reaches the target and the task is successfully completed.
    \item \textbf{False Positive}: The robot stops near a wrongly detected target, leading to task failure.
    \item \textbf{Stepout Feasible}: The target remains theoretically reachable, but the robot exceeds the step budget before completing the task.
    \item \textbf{No Frontier}: The robot runs out of valid traversable frontiers to explore before reaching the target.
\end{itemize}

\begin{figure*}[t]
		\centering
	    \includegraphics[width=0.95\textwidth]{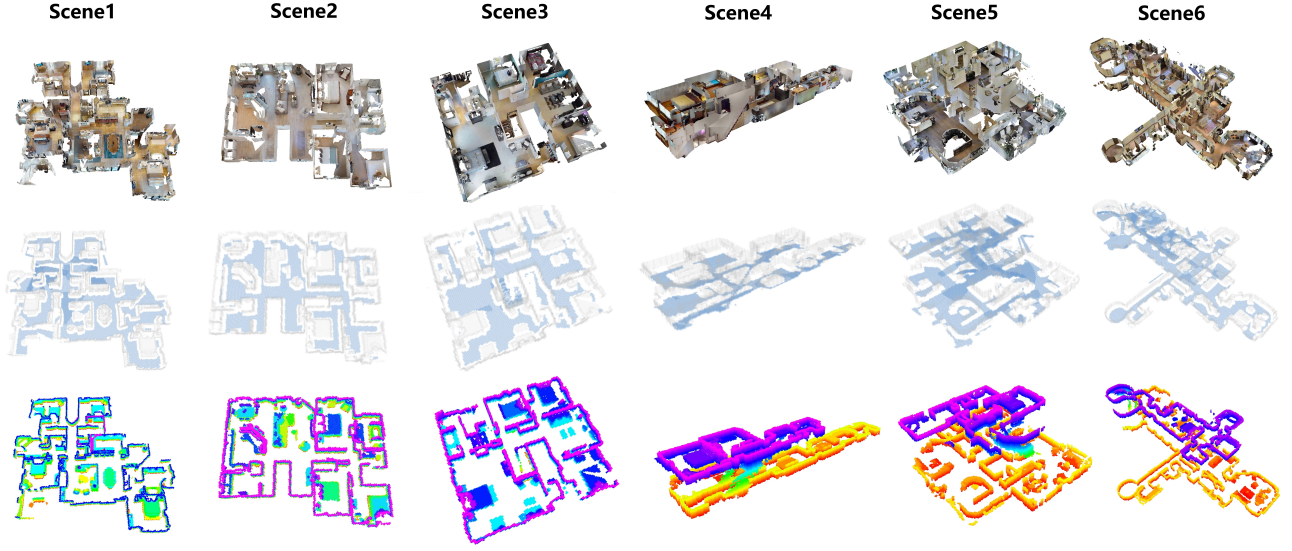}
	    \captionsetup{font=footnotesize}
		\caption{Qualitative mapping results in six representative indoor scenes.
Each column shows the original scene geometry, the reconstructed traversability map, and the corresponding occupancy map from top to bottom. 
Scenes 1--3 are single-floor scenes, and Scenes 4--6 are multi-floor scenes.
The traversability map captures robot-reachable support surfaces, while the occupancy map represents surrounding structures and obstacles, together supporting traversable frontier extraction and 3-D planning.}
\label{fig:map_layers}
\end{figure*}

As shown in Fig.~\ref{fig:failure_modes}, the proposed system achieves a success rate of $70\%$ in the complete evaluation setting. Among the remaining failure cases, \textit{False Positive} is the dominant factor, accounting for $22\%$ of all trials. 
This is mainly caused by the multi-frame inconsistency of the detector, which leads to unstable semantic observations during exploration. 
With the rapid progress of vision foundation models, this limitation is expected to be substantially alleviated in the near future.
Another major cause is annotation error: the agent may find the target, but the task still fails because the object is unlabeled in the dataset, as also observed in \cite{zhang2025apexnav,gong2026ascent}. 
In contrast, exploration-related failures are less frequent. \textit{Stepout Feasible} accounts for $6\%$, indicating that the target can still be reached in principle, but the robot fails to complete the task within the given step budget. These cases usually occur in large-scale environments or scenes requiring long detours. \textit{No Frontier} only accounts for $2\%$, suggesting that the traversable-frontier extraction and update strategy can maintain sufficient exploration candidates in most scenes.

The cross-floor setting is significantly more challenging. 
Compared with the complete setting, \textit{False Positive} remains the largest failure source, increasing to $26\%$. 
This indicates that most failures are still caused by annotation errors and the limited capability of the detector, rather than the exploration strategy itself. 
Importantly, \textit{Stepout Feasible} rises from $6\%$ to $18\%$. This shows that long-distance targets are difficult to locate using a greedy exploration strategy under a limited step budget. The robot may be attracted to locally promising regions, while the true target requires longer-range reasoning across floors and partially explored areas. Nevertheless, our cross-floor transition strategy is reasonable, since aggressively moving to another floor too early can introduce more revisits and unnecessary detours before sufficient evidence is collected on the current floor. \textit{Frontier Exhausted} also increases from $2\%$ to $7\%$.
This is mainly caused by discontinuous support surfaces, narrow stair transitions, and occluded regions in cross-floor scenes. Moreover, incomplete reconstruction and holes in the habitat simulator lead to inaccurate traversability estimation, which breaks frontier connectivity and causes the robot to exhaust
the valid frontiers before finding the correct transition.

\begin{figure}[t]
	\centering
	\includegraphics[width=1.0\linewidth]{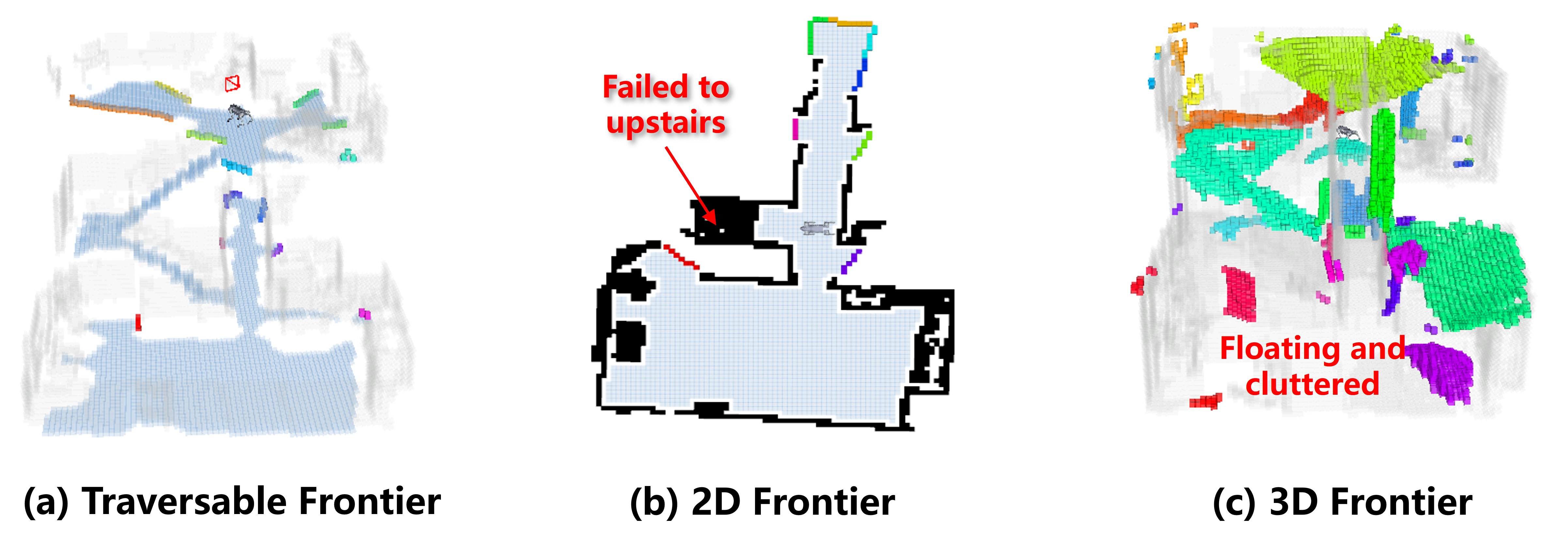}
    \captionsetup{font=footnotesize} 
	\caption{Qualitative comparison of frontier representations.
(a) TravExplorer extracts frontiers on connected traversable support surfaces, yielding reachable exploration goals.
(b) 2-D frontiers collapse vertical structures and miss the upstairs transition.
(c) Unconstrained 3-D frontiers produce floating and cluttered candidates that are not executable for a ground robot.
}
	\label{fig:frontier_ablation}
\end{figure}

\subsection{Qualitative Results}
\textbf{3-D Mapping Performance.}
We first evaluate the 3-D mapping performance of TravExplorer in diverse indoor layouts. 
Across 6 representative reconstructions, including both single-floor scenes and multi-floor buildings, the system consistently constructs multiple maps required for 3-D exploration, including traversability and occupancy maps, as shown in Fig.~\ref{fig:map_layers}.
The textured map is only used for scene visualization and is not involved in the planning pipeline.
In single-floor scenes, the traversability layer separates walkable regions from walls, furniture, and other non-supporting structures.
In multi-floor scenes, stairs, landings, and vertically overlapping rooms are preserved in a unified 3-D metric frame rather than being collapsed into a top-down grid.
This representation provides executable map memory for subsequent exploration, because frontiers and semantic values are attached only to visible and robot-connected traversable regions.

\textbf{Ablation study of Traversable Frontier.}
Fig.~\ref{fig:frontier_ablation} compares the proposed traversable-frontier representation with conventional 2-D and unconstrained 3-D frontiers.
The 2-D frontier representation projects the environment to a planar grid and therefore loses vertical connectivity; this limitation is consistent with the cross-floor result of 2-D-frontier baselines such as VLFM, which achieves only $0.4\%$ SR.
In contrast, unconstrained 3-D frontiers preserve height information but generate many floating or obstacle-adjacent candidates that cannot be used as executable viewpoints for a ground robot.
TravExplorer extracts frontier clusters directly from connected traversable support surfaces and filters candidates according to obstacle clearance and reachability. 
The design enables the frontier-based exploration to remain valid in both ordinary planar exploration and cross-floor search. 
The robot can continuously explore along reachable support
surfaces, identify stair-related transitions, and maintain generalization across different indoor layouts.

\begin{figure}[t]
	\centering
	\includegraphics[width=1.0\linewidth]{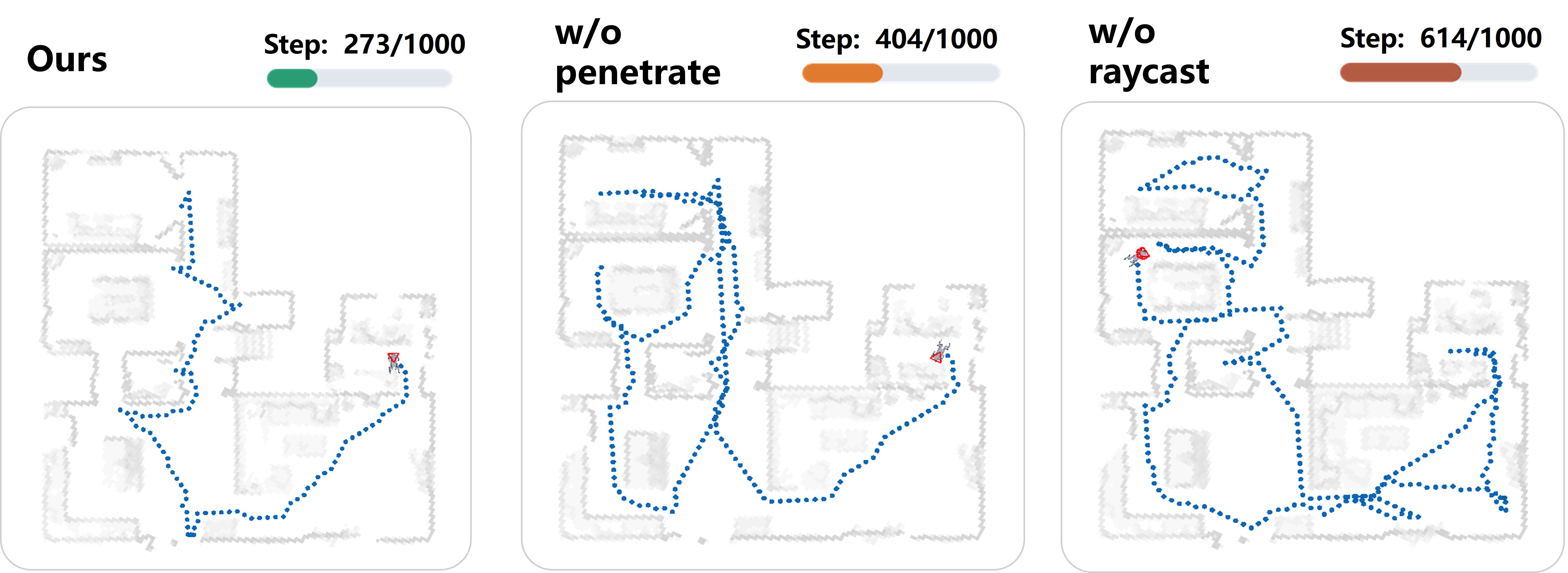}
    \captionsetup{font=footnotesize} 
	\caption{Mapping ablation on exploration efficiency.
TravExplorer reaches the target in 273/1000 steps, while removing penetration-based traversability reasoning or raycast-based visibility checking increases the required exploration to 404/1000 and 614/1000 steps, respectively.
The comparison shows that reliable traversability reasoning and visibility-aware frontier updating reduce redundant search and backtracking during target-driven exploration.
}
	\label{fig:efficiency_ablation}
\end{figure}

\textbf{Ablation study of Adaptive Mapping.}
We examine why object-goal exploration cannot rely on semantic segmentation alone.
We ablate two key designs in the hybrid and adaptive mapping module:
penetration-based raycasting and raycast-based traversability mapping, as visualized in Fig.~\ref{fig:efficiency_ablation}. 
In a representative planar episode, the complete system reaches the target in $273/1000$ steps. 
Replacing penetration-based raycasting with non-penetrating raycasting increases the required exploration to $404/1000$ steps, indicating that penetration improves exploration efficiency for navigation. 
Since the ObjectNav task does not require complete, accurate scene reconstruction, the robot should avoid carefully scanning every room corner and instead focus on regions that are more relevant to the target.
Removing the 2-D raycast-based mapping and relying only on semantic segmentation further increases the required exploration to $614/1000$ steps. 
This degradation is mainly caused by missed and inconsistent semantic predictions, which induce repeated local revisits. 
These results show that semantic segmentation alone is insufficient, while the proposed hybrid mapping reduces redundant search by combining raycast-based geometric coverage with semantic traversability.

\subsection{Real-World Experiments}
We further conduct field experiments to validate the feasibility of the proposed exploration framework across four types of challenging real-world buildings. 

\begin{figure}[t]
	\centering
	\includegraphics[width=0.8\linewidth]{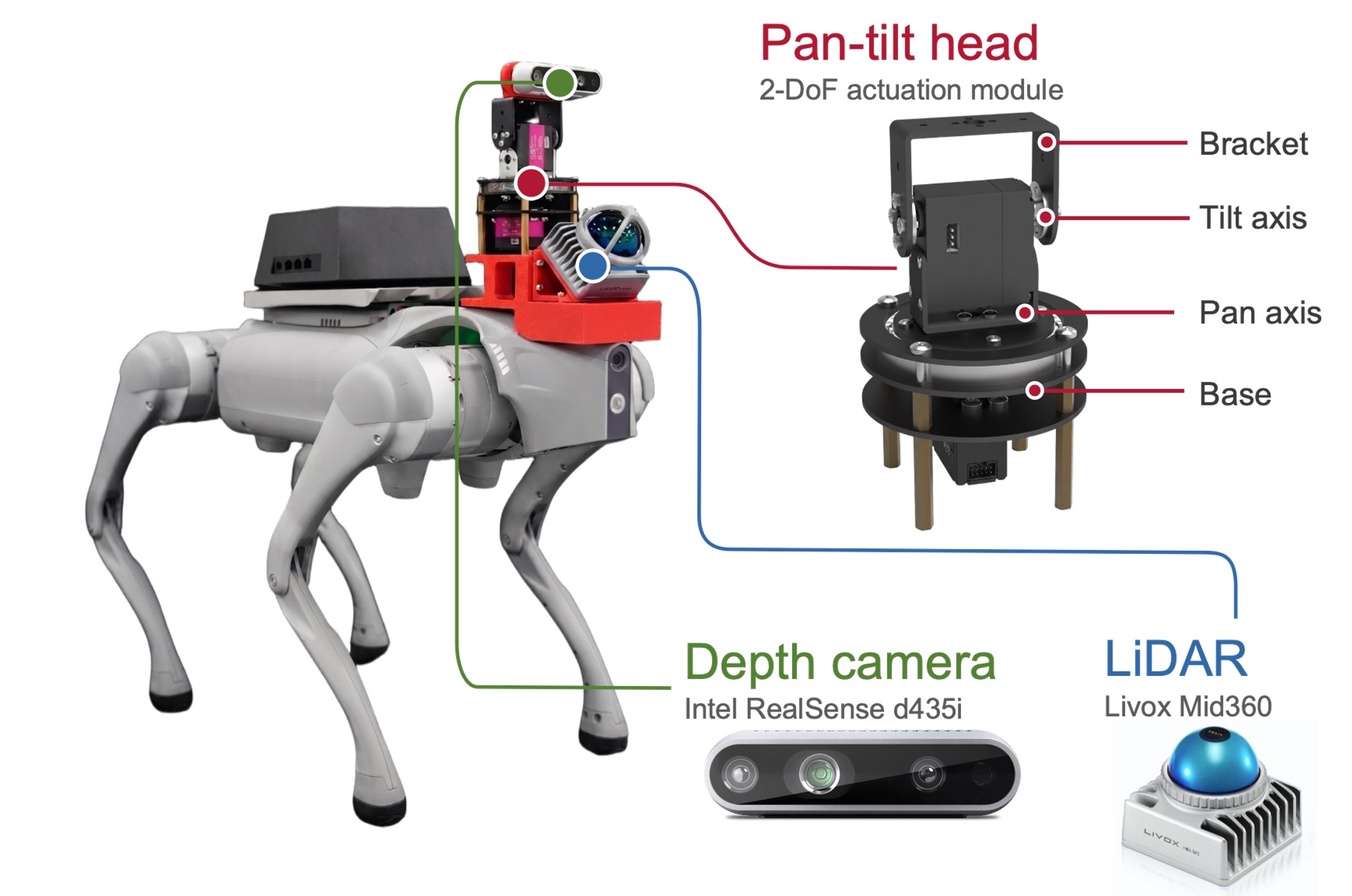}
    \captionsetup{font=footnotesize} 
	\caption{Hardware platform for real-world deployment.
TravExplorer is deployed on a Unitree Go2 quadruped robot equipped with an Intel RealSense D435i depth camera, a Livox Mid-360 LiDAR, and a custom 2-DoF pan--tilt perception head.
The LiDAR provides robust localization and geometric sensing, while the RGB-D camera supplies close-range visual and depth observations for occupancy, traversability, and semantic mapping.
The pan--tilt head actively adjusts the camera viewpoint to inspect floors, stairs, and occluded regions, supporting FOV-aware active perception and executable cross-floor exploration.
}
	\label{fig:platform}
\end{figure}

\textbf{Real-world robot setup.}
In real-world experiments, TravExplorer is deployed on the Unitree Go2 quadruped shown in Fig.~\ref{fig:platform}. 
The dynamic limits of the quadruped robot are set to $v_m=1.0\,\mathrm{m/s}$, $a_m=0.25\,\mathrm{m/s^2}$, and $\omega_m=1.0\,\mathrm{rad/s}$.
The built-in LiDAR of Go2 is removed and replaced with a Livox Mid-360 LiDAR to provide higher-quality point clouds. 
Our adapted FAST-LIO2 \cite{xu2022fast} provides high-frequency localization.
A forward-looking stereo camera, an Intel RealSense D435i~\cite{realsense_d435i}, is mounted on a 2-DoF pan-tilt head for active RGB-D perception.
The system runs under a ROS-based master-slave architecture. 
A workstation with an NVIDIA RTX 4090 GPU and an Intel i9-14900HX CPU serves as the ROS master, while the Go2 equipped with an NVIDIA Jetson Orin Nano acts as the slave platform. 
A portable Wi-Fi router is used for ROS communication between the laptop and the robot.

\begin{figure*}[t]
		\centering
	    \includegraphics[width=0.95\textwidth]{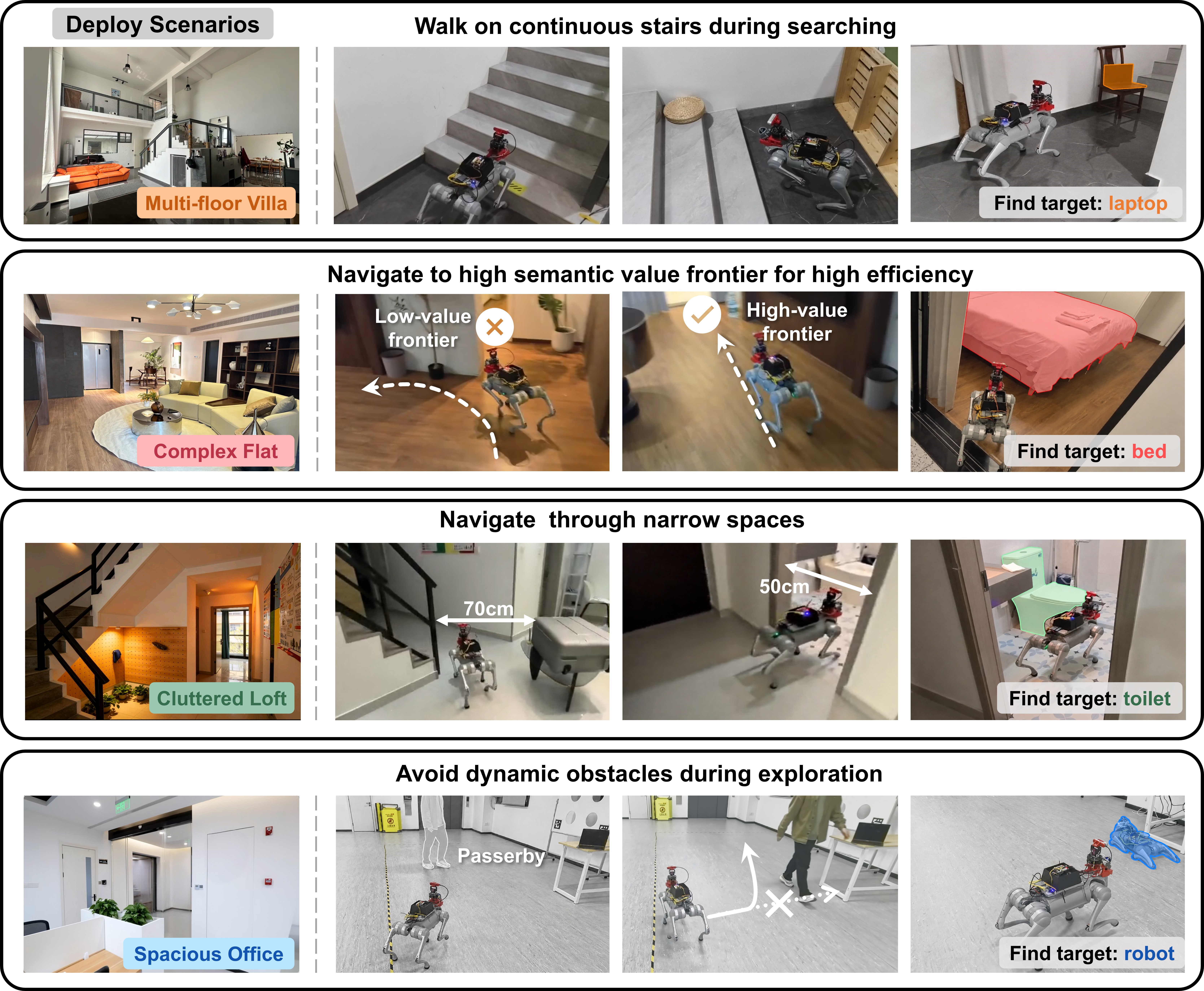}
	    \captionsetup{font=footnotesize}
		\caption{Real-world deployment results of TravExplorer in four representative indoor environments.
The first row shows deployment in a multi-floor villa, where the robot continuously ascends stairs during target search and reaches an executable viewpoint for the queried target, \textit{laptop}.
The second row shows a complex flat, where TravExplorer avoids a low-value frontier and selects a high-semantic-value frontier, enabling efficient navigation toward the queried target, \textit{bed}.
The third row shows a cluttered loft-like environment, where the robot navigates through narrow passages of approximately 70~cm and 50~cm width while maintaining safe local mapping and obstacle avoidance, and finally searches for the target, \textit{toilet}.
The fourth row shows a spacious office area with dynamic pedestrians, where TravExplorer rapidly replans collision-free local paths around moving obstacles while maintaining the semantic search objective for the queried target, \textit{robot}.
These deployments demonstrate the real-world capability of TravExplorer in cross-floor traversability reasoning, semantic frontier selection, narrow-space navigation, and dynamic obstacle avoidance.
}
\label{fig:real_world}
\end{figure*}

\textbf{Real-world results.}
We further conducted real-world experiments to validate the deployability of TravExplorer in challenging indoor buildings. 
The experiments were performed in four representative environments, as shown in Fig.~\ref{fig:real_world}: a multi-floor villa, a complex flat, a cluttered loft, and a spacious office. 
These scenes cover different spatial scales, floor structures, passage widths, occlusion levels, and dynamic disturbances, including approximately $300~\mathrm{m}^2$ over two floors in the villa, $165~\mathrm{m}^2$ on a single floor in the flat, $188~\mathrm{m}^2$ in the loft-like space, and $450~\mathrm{m}^2$ over two floors in the office. 
All trials were started without pre-scanned maps, manually specified intermediate waypoints, or human guidance during execution.

Across the four environments, we conducted 50 physical trials. 
TravExplorer successfully completed 32 trials, corresponding to a real-world success rate of $64.0\%$, as shown in Table~\ref{tab:real_world_results}. 
The start locations and target queries were varied within each environment. The queried targets included beds, trash bins, plants, robot-like objects, and other user-specified indoor objects. 
A trial was counted as successful only if the robot reached a safe pose near the queried target, without manual teleoperation, emergency stop and within a timeout of 300s. 

In the multi-floor villa, continuous stair-ascent episodes demonstrate that the proposed 3-D traversability map can represent stairs, landings, and upper-floor surfaces as connected reachable support structures. 
In the complex flat, the robot used the semantic value map to downweight low-value frontiers and prioritize high-value semantic frontiers related to the queried target, thereby improving exploration efficiency in a multi-room layout. 
In the cluttered loft, TravExplorer showed accurate local mapping, obstacle avoidance, and narrow-passage navigation under dense furniture and partial visual occlusions. 
Even with incomplete observations, the robot could actively perceive and recognize targets. 
In the spacious office, TravExplorer rapidly replanned local collision-free paths around moving pedestrians while maintaining the current semantic search objective or frontier intention.
These results validate the effectiveness and deployability of TravExplorer for both single-floor and cross-floor embodied exploration in real indoor environments.

\begin{table}[t]
\centering
\caption{
Quantitative summary of real-world results.
}
\label{tab:real_world_results}
\scriptsize
\setlength{\tabcolsep}{1.9pt}
\begin{tabular}{l c c c c c c c c}
\toprule
\textbf{Env.} 
& \textbf{Trials} 
& \textbf{Succ.} 
& \textbf{SR} 
& \textbf{AT} 
& \textbf{SPT} 
& \textbf{Manual} 
& \textbf{E-stop} 
& \textbf{Timeout} \\
\midrule
Villa  
& 7  
& 4  
& 57.1\% 
& 203.0~s 
& 18.5\% 
& 28.6\% 
& -- 
& 14.3\% \\

Flat   
& 5  
& 3  
& 60.0\% 
& 46.0~s 
& 50.8\% 
& 20.0\% 
& 20.0\% 
& -- \\

Loft   
& 18 
& 12 
& 66.7\% 
& 57.3~s 
& 53.9\% 
& 5.6\% 
& 16.7\% 
& 11.1\% \\

Office 
& 20 
& 13 
& 65.0\% 
& 101.8~s 
& 42.9\% 
& 5.0\% 
& 15.0\% 
& 15.0\% \\
\midrule
\textbf{Overall} 
& \textbf{50} 
& \textbf{32} 
& \textbf{64.0\%} 
& \textbf{92.5~s} 
& \textbf{44.3\%} 
& \textbf{10.0\%} 
& \textbf{14.0\%} 
& \textbf{12.0\%} \\
\bottomrule
\end{tabular}
\vspace{2pt}

\parbox{\columnwidth}{
}
\end{table}

\section{Conclusion}

This article presented \textit{TravExplorer}, a traversability-aware embodied exploration framework for zero-shot ObjectNav in cross-floor indoor buildings. 
The proposed framework addresses the coupling between semantic target search and physically executable navigation by maintaining a unified 3-D traversability map, where floors, stairs, and landings are represented as connected support surfaces while obstacles are kept separate. 
Based on this representation, traversable frontiers are incrementally extracted on robot-reachable surfaces, FOV-limited stair observations are completed through active perception, and target-related evidence is accumulated through probabilistic instance memory and spatial value projection. 
A hierarchical planner further integrates target-aware frontier touring, stair-landmark selection, foothold-guided 3-D graph search, an execute-review mechanism, and vertically constrained local trajectory optimization to convert semantic exploration decisions into executable robot motions.

Extensive simulation and real-world experiments validated the effectiveness of the proposed framework. 
On HM3D and MP3D, TravExplorer consistently improves both success rate and SPL over representative learning-based, zero-shot, and floor-aware ObjectNav baselines, with the largest gains observed in multi-floor episodes that require explicit stair and landing connectivity. 
Real-world deployment on a Unitree Go2 quadruped further demonstrates open-vocabulary target search in single-floor and cross-floor buildings without prior maps or human waypoint guidance, while handling semantic frontier selection, narrow-passage navigation, cross-floor traversal, and local replanning around dynamic obstacles. 
Future work will focus on extending TravExplorer from a rule-based system
toward a general agentic framework, where large vision-language models are
used for high-level reasoning, and prediction over unobserved regions.

\bibliographystyle{IEEEtran}
\bibliography{IEEEabrv, example}  

\end{document}